\documentclass[journal]{IEEEtran}

\usepackage{cite}

\ifCLASSINFOpdf
\else
\fi
\usepackage{amssymb}
\usepackage[utf8]{inputenc}
\usepackage{amsmath,graphicx}
\usepackage{xspace}
\usepackage{booktabs}
\usepackage{color}
\usepackage{multirow}
\usepackage[bookmarks,bookmarksnumbered,colorlinks=true,linktocpage]{hyperref}
\usepackage{url}

\begin{document}
%
\title{Semantic Diversity versus Visual Diversity \\ in Visual Dictionaries}
%
%
%

\author{Ot\'{a}vio~A.~B.~Penatti,
        Sandra~Avila,~\IEEEmembership{Member,~IEEE,}
        Eduardo~Valle,
        Ricardo~da~S.~Torres,~\IEEEmembership{Member,~IEEE}
\thanks{O. A. B. Penatti is with the SAMSUNG Research Institute, Campinas, Brazil (e-mail: o.penatti@samsung.com).}
\thanks{S. Avila and E. Valle are with the School of Electrical and Computing Engineering, University of Campinas (Unicamp), Campinas, Brazil.} 
\thanks{R. da S. Torres is with the Institute of Computing, University of Campinas (Unicamp), Campinas, Brazil.}
}

\maketitle

\begin{abstract}

Visual dictionaries are a critical component for image classification/retrieval systems based on the bag-of-visual-words (BoVW) model.
Dictionaries are usually learned without supervision from a training set of images sampled from the collection of interest.
However, for large, general-purpose, dynamic image collections (e.g., the Web), obtaining a representative sample in terms of semantic concepts is not straightforward.
In this paper, we evaluate the impact of semantics in the dictionary quality, aiming at verifying the importance of semantic diversity in relation visual diversity for visual dictionaries.
In the experiments, we vary the amount of classes used for creating the dictionary and then compute different BoVW descriptors, using multiple codebook sizes and different coding and pooling methods (standard BoVW and Fisher Vectors).
Results for image classification show that as visual dictionaries are based on low-level visual appearances, visual diversity is more important than semantic diversity.
Our conclusions open the opportunity to alleviate the burden in generating visual dictionaries as we need only a visually diverse set of images instead of the whole collection to create a good dictionary.

\end{abstract}


%
\IEEEpeerreviewmaketitle

\section{Introduction}

    Among the most effective representations for image classification, many are based on visual dictionaries, in the so-called Bag of (Visual) Words (BoW or BoVW).
    The BoVW model brings to Multimedia Information Retrieval many intuitions from Textual Information Retrieval~\cite{SivicVideoGoogle2003}, mainly the idea of using occurrence statistics on the contents of the documents (words or local patches), while ignoring the document structure (phrases or geometry).

    The first challenge in translating the BoVW model from textual documents to images is that the concept of a ``visual word'' is much less straightforward than that of a textual word.
    In order to create such ``words,'' one employs a visual dictionary (or codebook), which organizes image local patches into groups according to their appearance.
    This description fits the overwhelming majority of works in the BoVW literature, in which the ``visual words'' are exclusively appearance-based, and do not incorporate any semantic information (e.g., from class labels).
    This is in stark contrast to textual words, which are semantically rich.
    Some works incorporate semantic information from the class labels into the visual dictionary~\cite{Perronnin2006ECCV,Lazebnik2009PAMI,BoureauMidLevelCVPR2010}, but the benefit brought by those schemes is, so far, small compared to their cost, especially for large-scale collections. 

    The BoVW model is a three-tiered model based on: (1)~the extraction of low-level local features from the images, (2)~the aggregation of those local features into mid-level BoVW feature vectors (guided by the visual dictionary), (3)~the use of the BoVW feature vectors as an input for a classifier (often Support Vector Machines, or SVM).
    The most traditional way to create the visual dictionary is to sample low-level local features from a learning set of images, and to employ unsupervised learning (often $k$-means clustering) to find the $k$ vectors that best represent the learning sample.


    Aiming at representativeness, the dictionary is almost always learned on a training set sampled from the same collection that will be used for training and testing the classifier.
    In the closed datasets~\cite{TorralbaBiasDatasetCVPR2011} popularly used in the literature, the amount of images is fixed, therefore no new content is added after the dictionary is created.
    However, in a large-scale dynamic scenario, like the Web, images are constantly inserted and deleted.
    Updating the visual dictionary would imply updating all the BoVW feature vectors, which is unfeasible.

    It might be possible to solve this problem if we remind that the term ``visual dictionary'' is somewhat a misnomer, because it is not concerned with semantic information.
    The visual dictionaries we are interested in this work ignore the class labels and consider only the appearance-based low-level features.
    Provided that the selected sample represents well enough that low-level feature space (i.e., being diverse in terms of appearances), the dictionary obtained will be sufficiently accurate, even if it does not represent all semantic variability.

    We have seen recently an exponential growth of deep learning, mainly convolutional (neural) networks or ConvNets, for visual recognition~\cite{krizhevsky2012imagenet,OverFeat2014,OverFeat2014Generality}.
    However, some recent works~\cite{Perronnin2015CVPRFisherAndCNN,Klein2015CVPRFisherAndCNN} are joining both worlds: ConvNets and techniques based on visual dictionaries, like Fisher Vectors.
    Therefore, visual codebooks still tend to remain in the spotlight for some years.

    This paper evaluates the hypothesis that the quality of a visual dictionary depends only on the visual diversity of the samples used.
    The experiments evaluate the impact of the semantic diversity in relation to the visual diversity, varying the amount of classes used as basis for the dictionary creation.
    %
    Results point to the importance of visual variability regardless of semantics.
    We show that dictionaries created on few classes (i.e., low semantic variability) are able to produce good image representations that are as good as representations that use dictionaries based on the whole set of classes.
    To the best of our knowledge, there is no similar objective evaluation of visual dictionaries in the literature.


\section{Related work}
\label{related_work}

    We start this section with a review of the existing art on dataset dependency of dictionaries.
    Then, we detail the representation based on visual dictionaries.
    This work focuses on visual dictionaries created by unsupervised learning, since there are several challenges in scaling-up supervised dictionaries. 
    However, at the end of this section, we also briefly discuss how supervised dictionaries contrast to our experiments.

\subsection{Dictionaries and dataset dependency}
\label{dictionaries_dataset}

    There is scarce literature on the impact of the sampling choice on the quality of visual dictionaries.
    According to Perronnin and Dance~\cite{PerronninFisherCVPR2007}, dictionaries trained on one set of categories can be applied to another set without any significant loss in performance.
    They remark that the Fisher kernels are fairly insensitive to the quality of the generative models and therefore that the same visual dictionary can be used for different category sets.
    However, they do not evaluate the impact of semantics in the dictionary quality. 

    To tackle the problem of dataset dependency for the visual dictionaries,
    Liao et al.~\cite{Liao2013CrossCodebook} transform BoVW vectors derived from one visual dictionary to make them compatible with another dictionary. 
    By solving least squares, the authors showed that the cross-codebook method is comparable to the within-codebook one, when the two codebooks have the same size.

    Zhou et al.~\cite{ZhouScalaQuantizationMM2012} highlight many of the problems that we discuss here: the problem of dataset dependency for the visual dictionaries, the high computational cost for creating dictionaries, and the problem of update inefficiency.
    Although they do not provide experiments for confirming that, they propose a scalar quantization which is independent of datasets.




\subsection{Image representations based on visual dictionaries}
\label{visual_dictionaries}

    The bag-of-visual-words (BoVW) model~\cite{SivicVideoGoogle2003,CsurkaBagECCV2004} depends on the visual dictionary (also called visual codebook or visual vocabulary), which is used as basis for encoding the low-level features.
    BoVW descriptors aim to preserve the discriminating power of local descriptions while pooling those features into a single feature vector~\cite{BoureauMidLevelCVPR2010}.

    The pipeline for computing a BoVW descriptor is divided into (i)~dictionary generation and (ii)~BoVW computation.
    The first step can be performed off-line, that is, it is based on the training images only.
    And the second step computes the BoVW using the dictionary created in the first step. 
    

    To generate the dictionary, one usually samples or clusters the low-level features of the images.
    Those can be based on interest-point detectors or on dense sampling in a regular grid~\cite{SandeEvaluatingTPAMI2010,JurieCreatingCodebooks2005,ChatfieldEvaluationEncondingBMVC2011}.
    From each of the points sampled from an image, a local feature vector is computed, with SIFT~\cite{LoweSIFT2004} being a usual choice.
    Those feature vectors are then either clustered (often based on $k$-means) or randomly sampled in order to obtain a number of vectors to compose the visual dictionary~\cite{JurieCreatingCodebooks2005,ViitaniemiCodebooks2008,PapaVisualDictOPFICIP2011}.
    This can be understood as a feature space quantization.


    After creating the dictionary, the BoVW vectors can be computed.
    For that, the original low-level feature vectors need to be coded according to the dictionary -- \textit{coding} step.
    This can be simply performed by assigning one or more of the visual words to each of the points in an image.
    Popular coding approaches are based on \textit{hard} or \textit{soft} assignment~\cite{LiuSoftICCV2011,GemertVisuaWordAmbiguityPAMI2010}, although, several possibilities exist~\cite{YuLCCNIPS2009,YangSCCVPR2009,WangLLCCVPR2010,OliveiraSSCICRA2012,AvilaBossaNovaCVIU2013}.




    The \textit{pooling} step takes place after the coding step.
    One can simply average the codes obtained over the entire image (average pooling), can consider the maximum visual word activation (max pooling~\cite{BoureauMidLevelCVPR2010,YangSCCVPR2009}), or can use several other existing pooling strategies~\cite{LazebnikBeyondBags2006,LiuSoftICCV2011,FengLpNormCVPR2011,PenattiWSAPR2013}





    Recently, Fisher Vectors~\cite{PerronninFisherCVPR2007,PerronninFisherECCV2010} have gained attention due to their high accuracy for image classification~\cite{ChatfieldEvaluationEncondingBMVC2011,SanchezFisherIJCV2013,Huang_2014_TPAMI}.
    Fisher Vectors extend BoVW by encoding the average first- and second-order differences between the descriptors and visual words.
    In~\cite{PerronninFisherECCV2010}, Perronnin et al. proposed modifications 
    over the original framework~\cite{PerronninFisherCVPR2007} and showed that they could boost the accuracy of image classifiers.
    A recent comparison of coding and pooling strategies is presented in~\cite{KoniuszComparisonMidLevelCVIU2013}.

\subsection{Supervised visual dictionaries}
\label{supervised_dictionaries}

    Many supervised approaches have been proposed to construct discriminative visual dictionaries that explicitly incorporate category-specific information~\cite{BoureauMidLevelCVPR2010,WinnObjectCategorization2005,MoosmannFastCodebooksNIPS2007,PerronninAdaptedVocabPAMI2008,MairalSupervised2008,LazebnikSupervisedCodebookPAMI2009}.
    %
    In~\cite{GohUnsupervisedSupervisedCodebookECCV2012}, the visual dictionary is trained in unsupervised and supervised learning phases.
    The visual dictionary may even be generated by manually labeling image patches with a semantic label~\cite{GemertSemanticCodebookCVPR2006,VogelSemanticCodebookIJCV2007,LiuSemanticCodebookCVPR2009}. 
    Other more sophisticated techniques have been adapted to learn the visual dictionary, such as restricted Boltzmann machines.



    The main problem of those supervised dictionaries is that we need to know beforehand -- when the dictionary is being created -- all the possible classes in the dataset.
    For dynamic environments, this is extremely challenging. 
    Another issue is the difficulty of finding enough labeled data that~be representative of very large collections.
    Finally, many supervised approaches for dictionary learning rely on very costly optimizations, for a small increase in accuracy. 
    For those reasons, 
    we focus on dictionaries created by unsupervised learning, which also represent the majority of works in the literature.


\section{Experimental setup}
\label{experiments_setup}




    We evaluated the impact of semantics in the dictionary quality, by varying the number of classes used as basis for the dictionary creation.
    We tested different codebook sizes and different coding and pooling strategies.
    It is important to evaluate those parameters as there could exist effects that appear only on specific configurations.

    We report the results using the PASCAL VOC 2007 dataset~\cite{PascalVOC2007}, 
    a challenging image classification benchmark that contains significant variability in terms of object size, orientation, pose, illumination, position and occlusion~\cite{TorralbaBiasDatasetCVPR2011}.
    This dataset consists of 20 visual object classes in realistic scenes, with a total of 9,963 images categorized in macro-classes and sub-classes as person (\emph{person}), animal (\emph{bird}, \emph{cat}, \emph{cow}, \emph{dog}, \emph{horse}, \emph{sheep}), vehicle (\emph{aeroplane}, \emph{bicycle}, \emph{boat}, \emph{bus}, \emph{car}, \emph{motorbike}, \emph{train}), and indoor objects (\emph{bottle}, \emph{chair}, \emph{dinning table}, \emph{potted plant}, \emph{sofa}, \emph{tv/monitor}). 
    Those images are split into three subsets: training (2,501 images), validation (2,510 images) and test (4,952 images).
    Our experimental results are obtained on `train+val'/test sets.

    Before extracting features, all images were resized to have at most 100 thousand of pixels (if larger), as proposed in~\cite{AkataPAMI2014}.
    As low-level features, we have extracted SIFT descriptors~\cite{VLFEAT} from 24 $\times$ 24 patches on dense regular grid every 4 pixels at 5 scales. 
    The dimensionality of SIFT was reduced from 128 to 64 by using principal component analysis (PCA).
    We have used one million descriptors randomly selected and uniformly divided among the training images (i.e., the same number of descriptors per image) as basis for PCA and also for dictionary generation.
    As the random selection could also impact the results, two sets of one million descriptors were used.

    We used two different coding/pooling methods and their visual dictionaries were created accordingly.
        For BoVW~\cite{SivicVideoGoogle2003} (obtained with hard assignment and average pooling),
        we apply $k$-means clustering with Euclidean distance over the sets of one million descriptors explained above.
        We have varied the number of visual words in 1,024, 2,048, 4,096, and 8,192.

        For Fisher Vectors~\cite{PerronninFisherECCV2010}, 
        the descriptor distribution is modeled using a Gaussian mixture model, whose parameters are also trained over the sets of one million descriptors just mentioned, using an expectation maximization algorithm.
        The dictionary sizes were 128, 256, 512, 1,024, and 2,048 words.


    We used a classification protocol (one-versus-all) with support vector machines (SVM) with the most appropriate configuration for each coding/pooling method:
    RBF kernel for BoVW and linear kernel for Fisher Vectors~\cite{JKernelMachines}.
    Grid search was performed to optimize SVM parameters. 
    The classification performance was measured by the average precision (AP).






    One of the most important parameters in the evaluation was the variation in the number of image classes used for dictionary generation.
    We used 18 different samples from the VOC 2007 dataset.
    The initial selection of samples was based on selecting individual classes, varying from 1, 2, 5, 10 to all the 20 classes.
    However, as VOC 2007 is multilabel (i.e., an image can belong to more than one class), when selecting points from images of class \textit{cat}, for instance, we probably also select points of class \textit{sofa}.
    Therefore, we show our results in terms of the real number of classes used to create each sample.
    We count the number of classes that appear in each sample, even if only 1 image of a class is used.
    We also computed a measure that considers the frequency of occurrence of classes, which we called \textit{semantic diversity}, better estimating the real amount of semantics in the samples.
    The semantic diversity \textit{S} is measured by the following equation:
    \begin{equation}
        S = \sum\limits^{N}_{i=1}\frac{C_i+C'{_i}}{C_i},
    \label{eq_sem_div}
    \end{equation}

    \noindent where $N$ is the number of classes selected, $C_i$ is the number of images in the class $i$ and $C'{_i}$ is the number of images in the class $i$ that also appear in other classes.
    A summary of the samples used is shown in Table~\ref{tab_info_partial_choice}.

\begin{table}[t!]
\centering
\begin{tiny}

\begin{tabular}{@{ }crrrl@{ }}
\toprule
id & \textit{S} & $N_{c}$ & $N_{i}$ & initial classes selected \\  \hline
1 & 1.49 & 18 & 421 & dog \\
2 & 1.56 & 18 & 713 & car \\
3 & 2.24 & 15 & 445 & chair \\ \hline
4 & 3.14 & 11 & 327 & cow, bus \\
5 & 3.24 & 12 & 561 & cat, sofa \\
6 & 3.57 & 13 & 530 & dining, bird \\
7 & 3.93 & 15 & 501 & tv, motorbike \\
8 & 4.11 & 17 & 731 & horse, chair \\ \hline
9 & 7.16 & 19 & 1560 & cat, bird, train, plant, dog \\
10 & 8.75 & 20 & 1638 & car, sofa, bottle, bicycle, train \\
11 & 8.82 & 20 & 1306 & motorbike, sheep, horse, aeroplane, chair \\
12 & 8.84 & 20 & 2477 & cow, boat, table, bus, person \\
13 & 9.78 & 18 & 2602 & bus, chair, bicycle, sofa, person \\ \hline
14 & 16.28 & 20 & 3478 & cat, horse, person, sheep, train, tv/monitor, bird, cow, sofa, motorbike \\
15 & 16.65 & 20 & 3358 & aeroplane, bird, plant, chair, horse, car, sofa, dog, boat, bicycle \\
16 & 16.87 & 20 & 2491 & horse, cow, bus, aeroplane, table, chair, boat, bicycle, dog, cat \\
17 & 16.88 & 20 & 3749 & tv/monitor, car, motorbike, bottle, cow, bird, train, plant, sheep, person \\
18 & 18.02 & 20 & 2366 & bus, cow, boat, table, aeroplane, bicycle, bird, train, dog, bottle \\ \bottomrule
\end{tabular}

\end{tiny}
  \caption{Summary of the samples used as basis for the dictionary creation in all of the experiments in this section. \textit{S} represents the semantic diversity computed by Equation~\ref{eq_sem_div}, $N_{c}$ is the number of classes that appear at least once in each sample and $N_{i}$ is the number of images in the sample.}
  \label{tab_info_partial_choice}
\end{table}



    It is important to highlight that the selection of classes (dictionary classes) is used only for creating the dictionary.
    The results (target classes) are reported always considering the whole VOC 2007 dataset.

    For analyzing the results, we took the logit of the average precisions in order to have measures that behave more linearly (i.e., that are more additive and homogeneous throughout the scale).
    Thus each average precision was transformed using the function: $logit(x) = -10 \times \log_{10} (\frac{x}{1-x})$.
    Then, in order to get the global effect of the increasing number of dictionary classes, we have aggregated the average precisions for all target classes.
    To make the average precision of different target classes commensurable, we have ``erased'' the effect of the target classes themselves by subtracting the average and dividing the standard deviation obtained for all logit values for that target class.
    This procedure, commonly employed in statistical meta-analyses, is necessary to obtain an aggregated effect because some target classes are much harder to classify than others.
    Let us call those values the standardized logits of the average precisions, which will be used in our analyzes.

\section{Results}
\label{experiments_semantic}

    Each experiment configuration explained in Section~\ref{experiments_setup} generated a set of 40 points (20 target classes and 2 runs, each one with a 1 million feature set).
    After plotting the standardized logits of the average precisions, we have computed the linear regression aiming at determining the variation in results when varying from low to high semantics (increasing number of dictionary classes).

\begin{figure}[t!]
     \begin{center}
      \scriptsize
        \begin{tabular}{@{}c@{ }c@{}}
            \includegraphics[width=0.45\columnwidth]{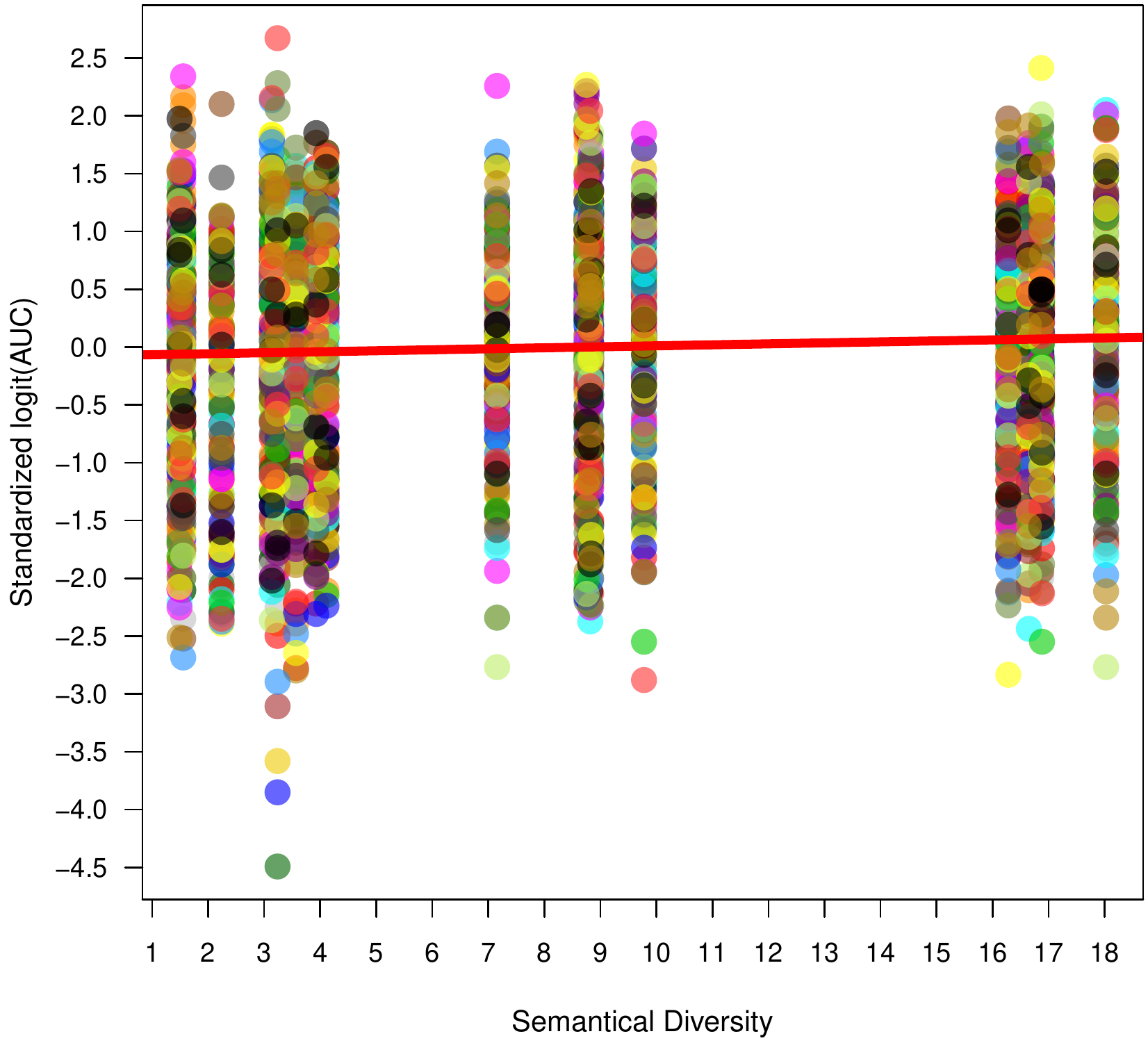} \hspace{0.5cm} &
            \includegraphics[width=0.45\columnwidth]{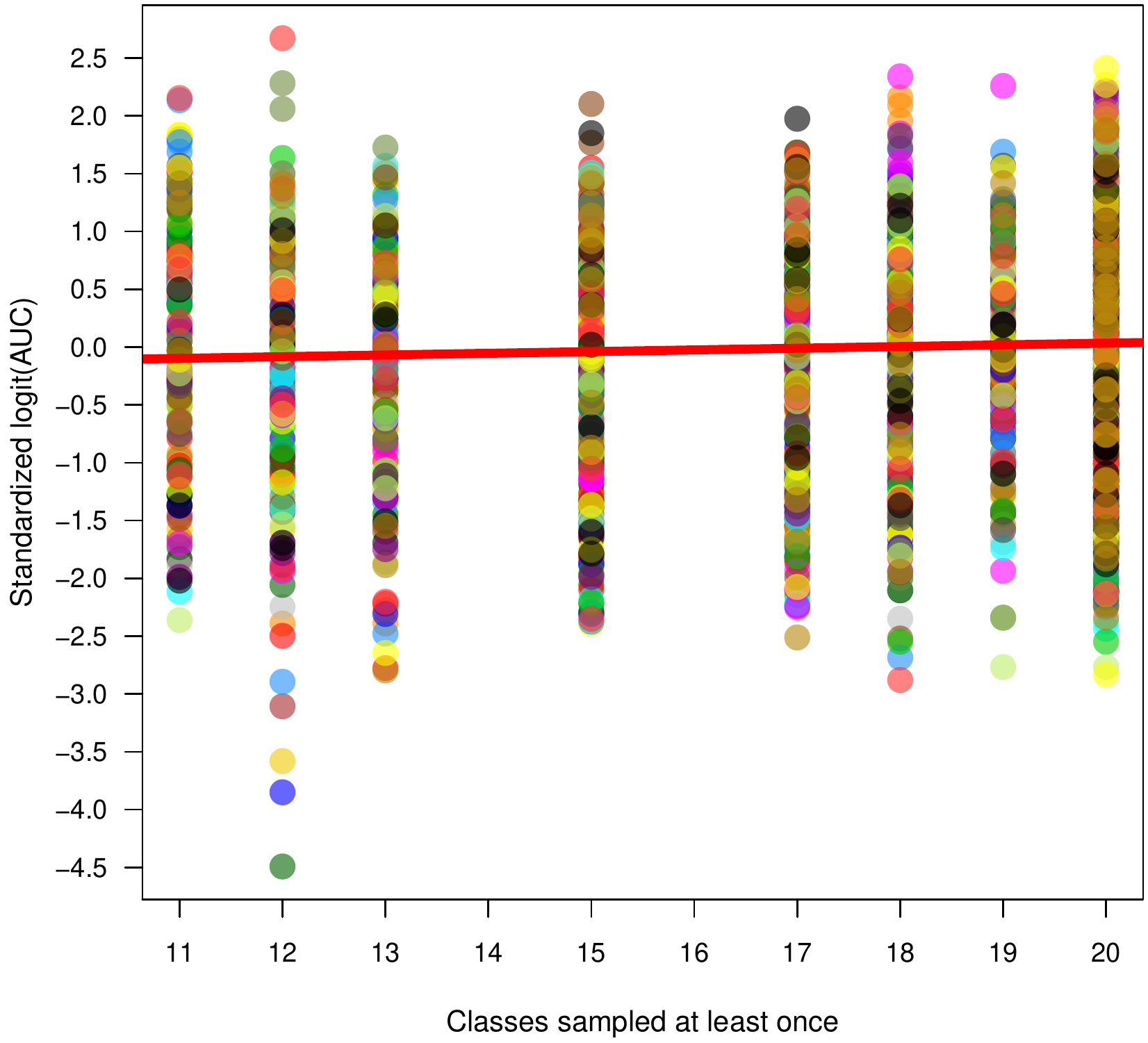} \\
            (a) $\alpha$ = 0.0085 (p-value=0.0071) & \hspace{0.1cm}
            (b) $\alpha$ = 0.0146 (p-value=0.0197) \\
        \end{tabular}
     \end{center}
     \caption{Standardized average precision of each class versus semantic variability considering all the runs of every BoVW configuration (all codebook sizes: 1024, 2048, 4096, 8192). The red line corresponds to the linear regression. The coefficient $\alpha$ (slope) is very small, showing that semantics has a very low impact on the codebook quality. Each of the 20 classes has a different dot color. In (a) the x-axis shows the semantic diversity, while in (b), the x-axis corresponds to the number of classes in each sample used to create the dictionary. Semantic diversity and number of classes are shown in Table~\ref{tab_info_partial_choice}.}
     \label{fig_bovw_plot_aps_vs_semantics_all}
\end{figure}

\begin{figure}[t!]
     \begin{center}
      \begin{tiny}
        \begin{tabular}{@{}c@{}c@{}c@{}c@{}}
            \includegraphics[width=0.25\columnwidth]{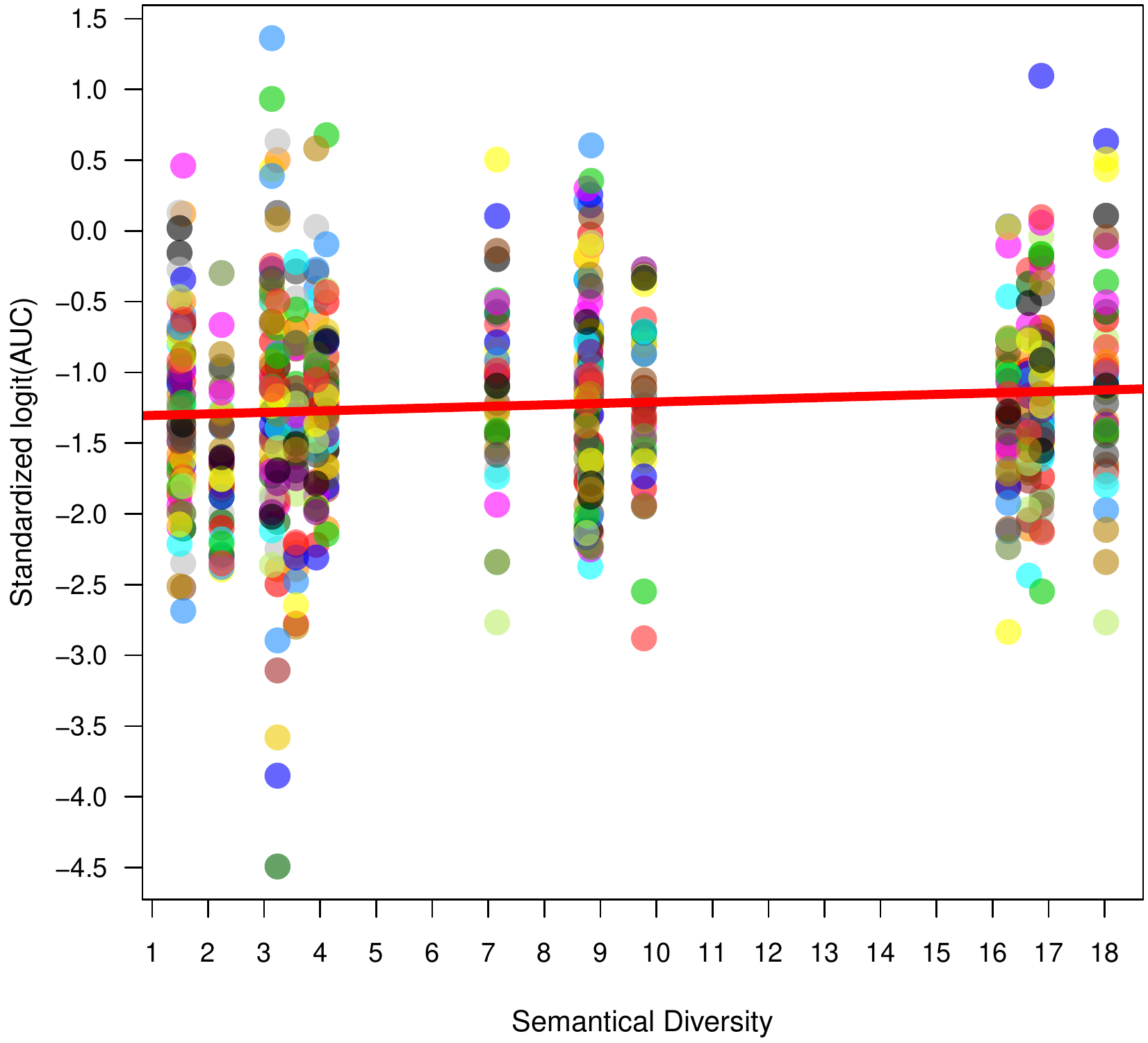} &
            \includegraphics[width=0.25\columnwidth]{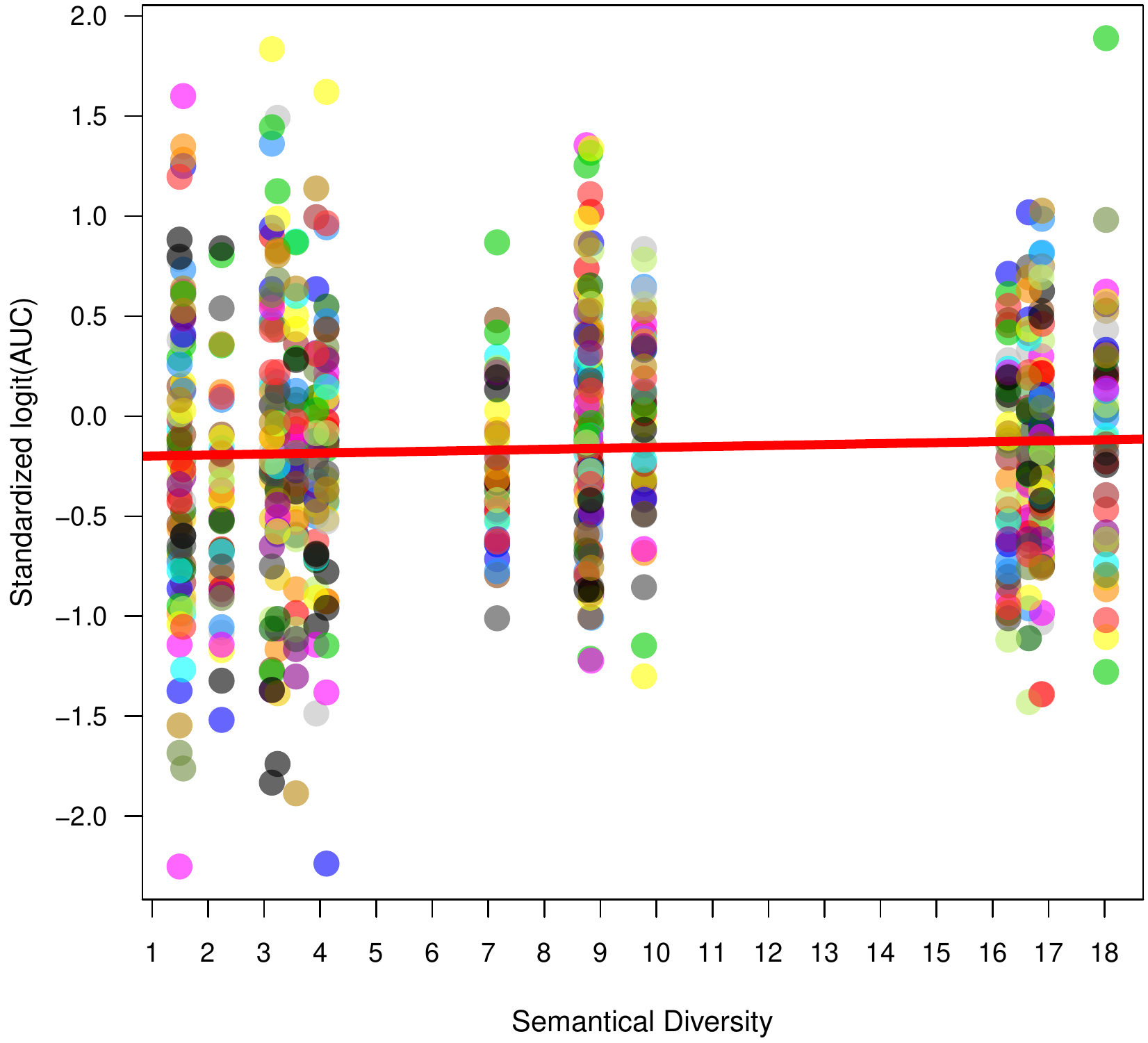} &
            \includegraphics[width=0.25\columnwidth]{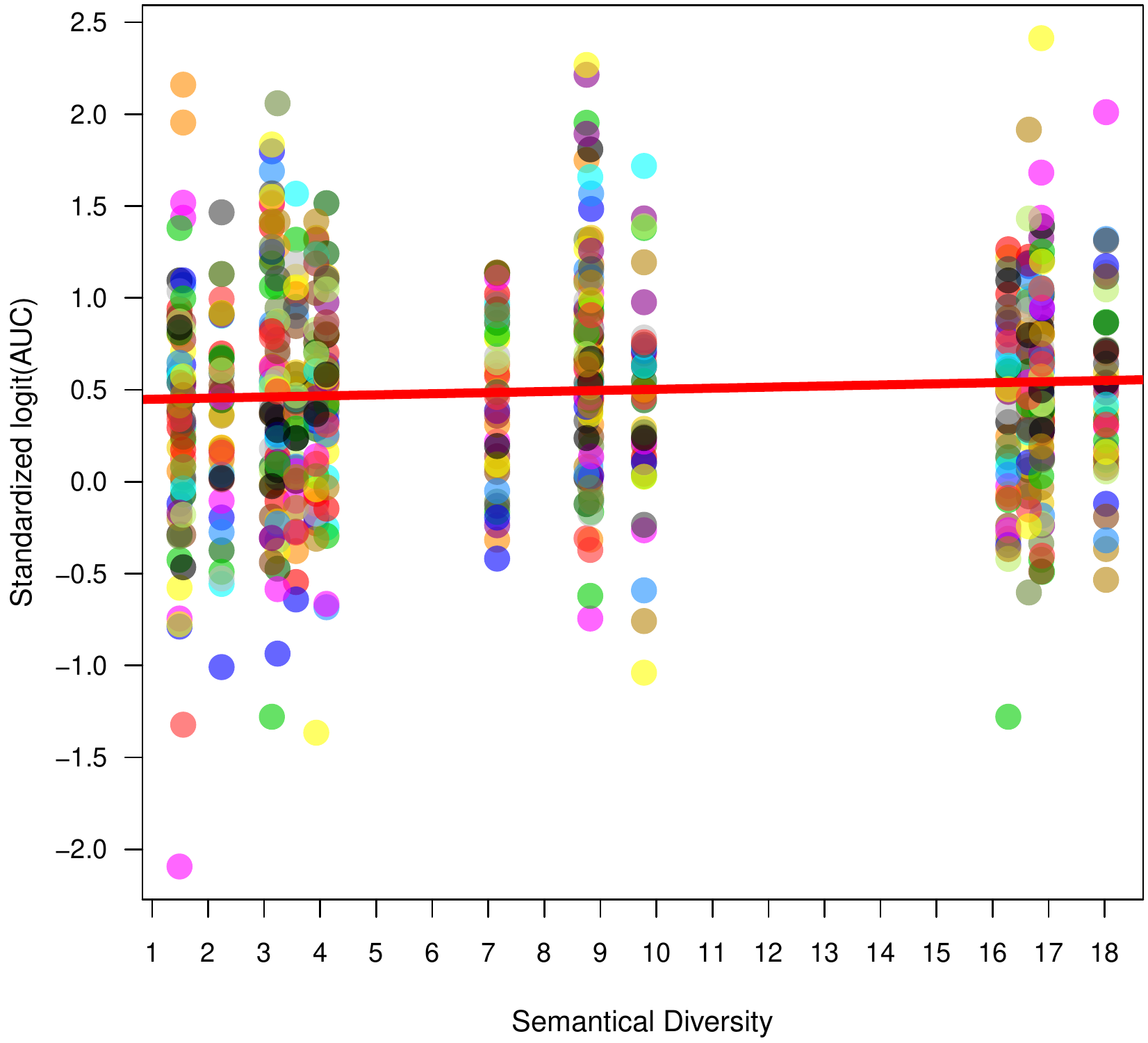} &
            \includegraphics[width=0.25\columnwidth]{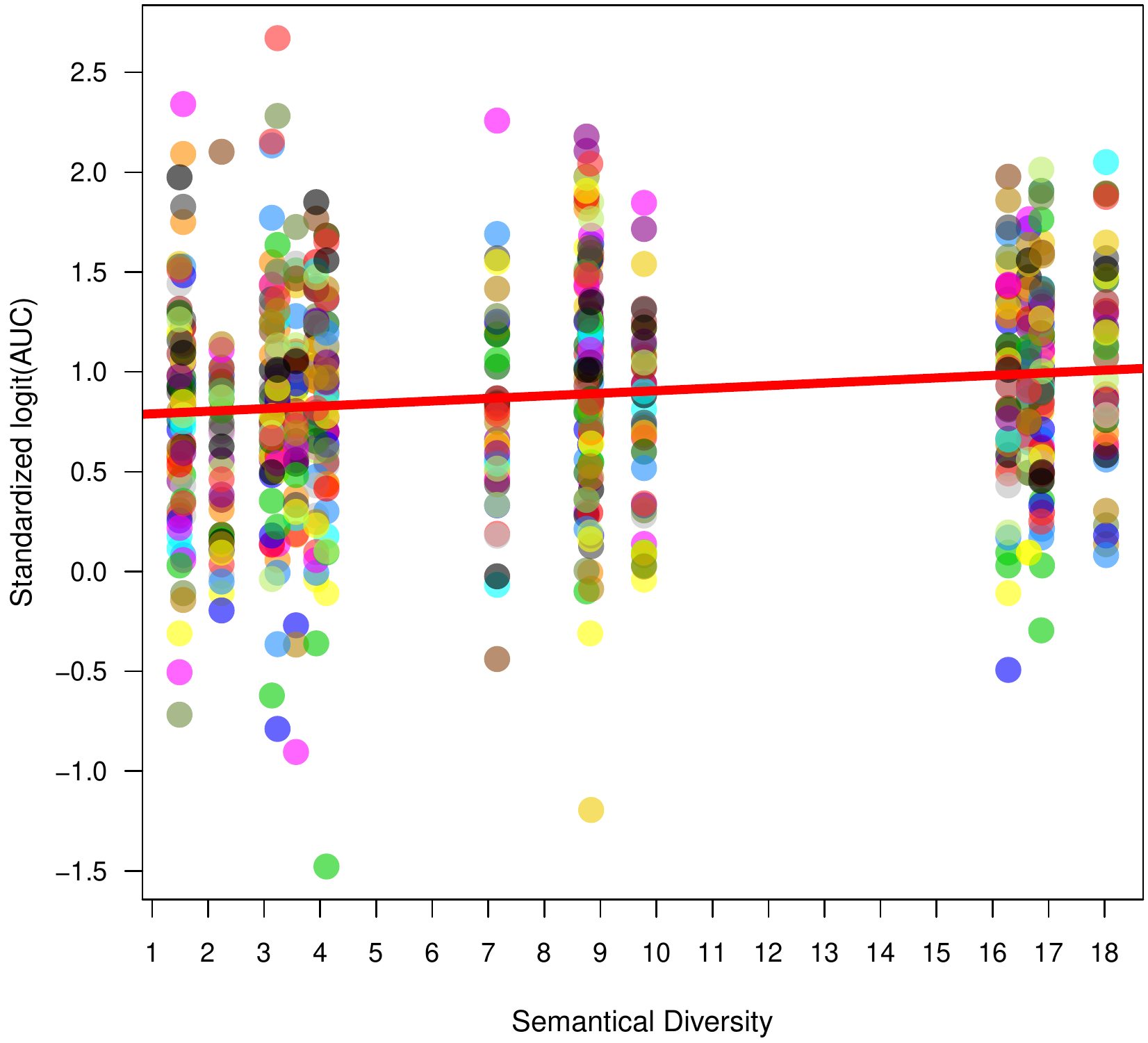} \\
            (a) 1024$\rightarrow$$\alpha$=0.011 &
            (b) 2048$\rightarrow$$\alpha$=0.005 &
            (c) 4096$\rightarrow$$\alpha$=0.006 &
            (d) 8192$\rightarrow$$\alpha$=0.013 \\
                (p-value=0.014) &
                (p-value=0.215) &
                (p-value=0.095) &
                (p-value=0.0002) \\

            \includegraphics[width=0.25\columnwidth]{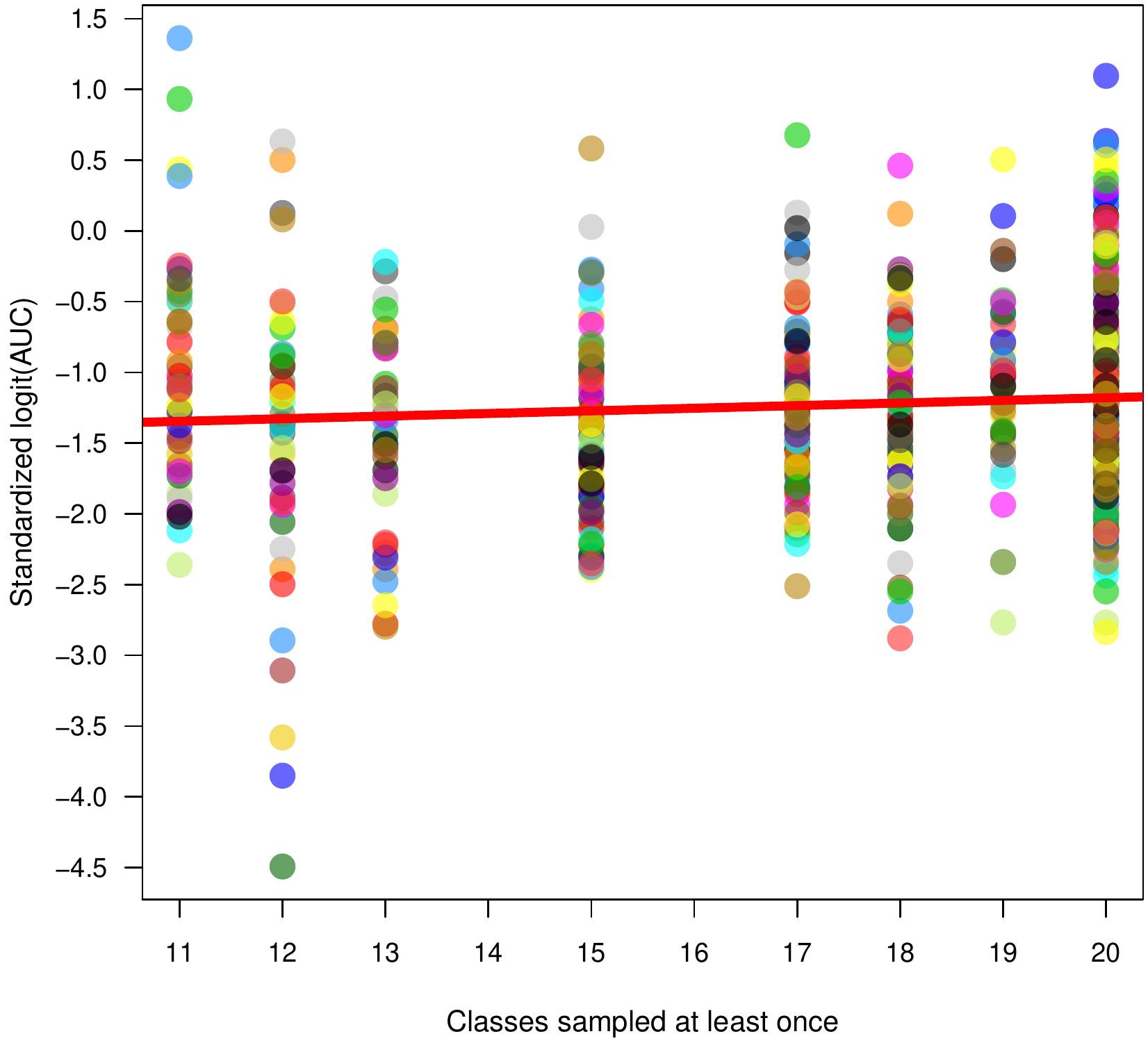} &
            \includegraphics[width=0.25\columnwidth]{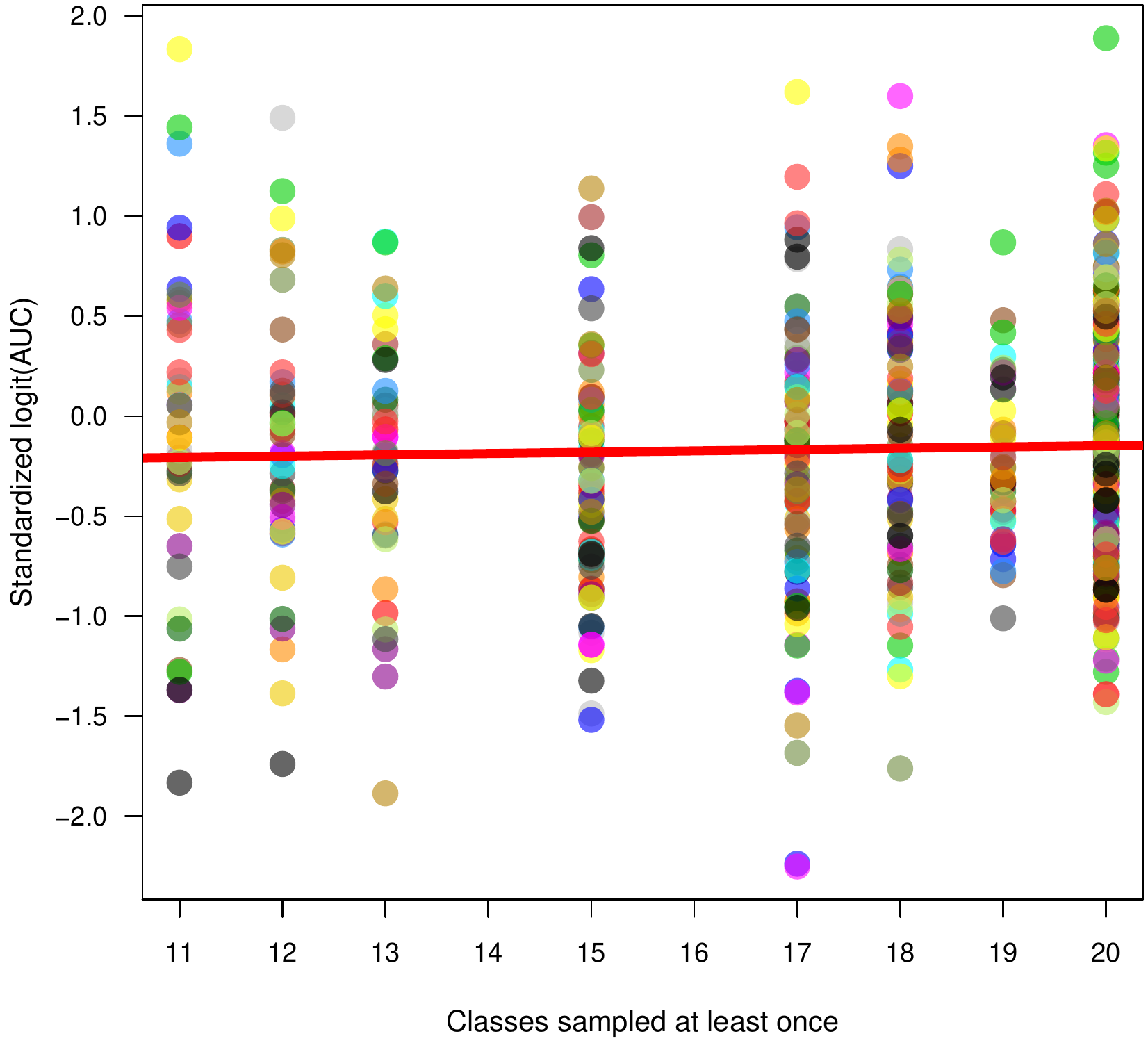} &
            \includegraphics[width=0.25\columnwidth]{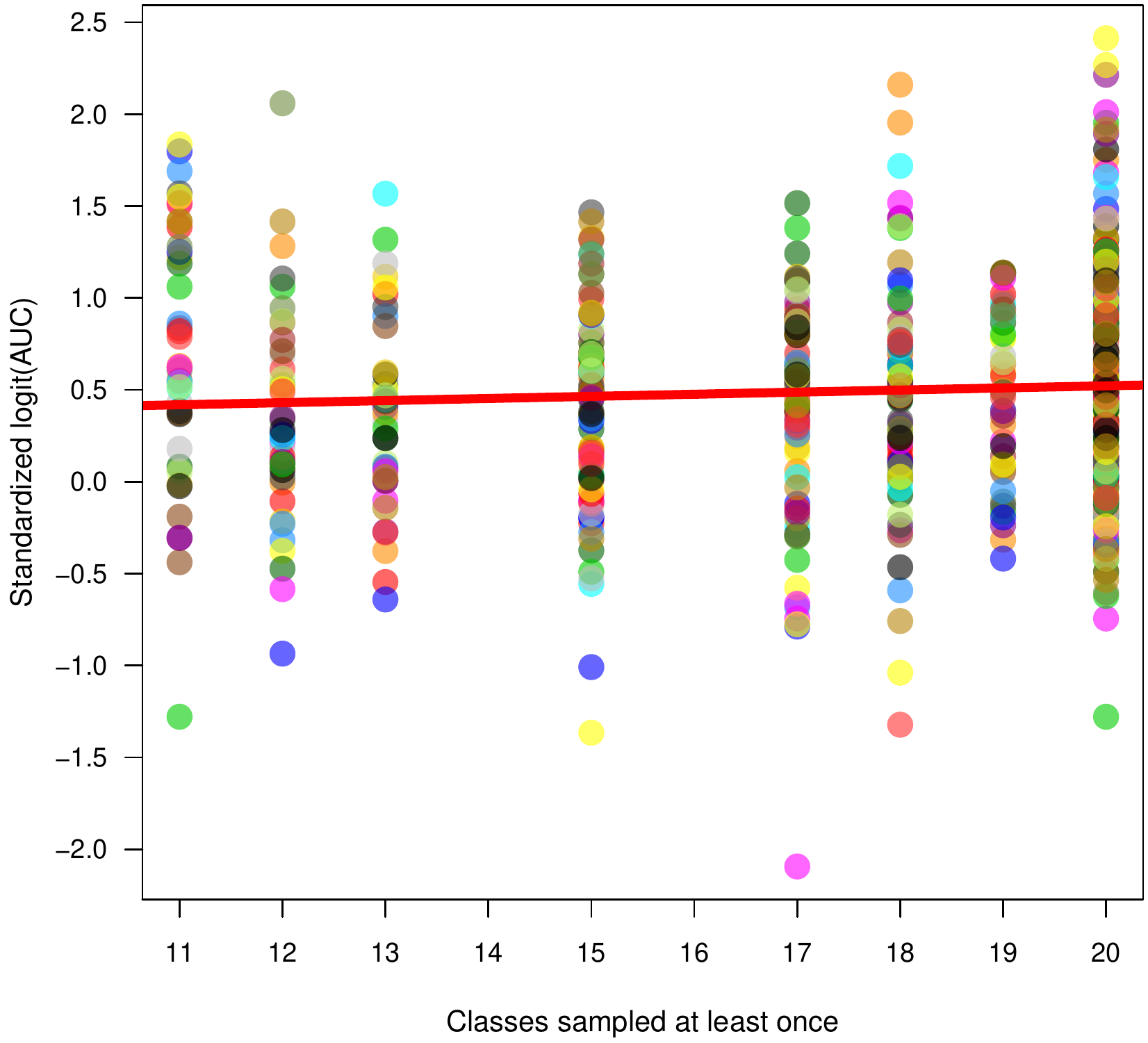} &
            \includegraphics[width=0.25\columnwidth]{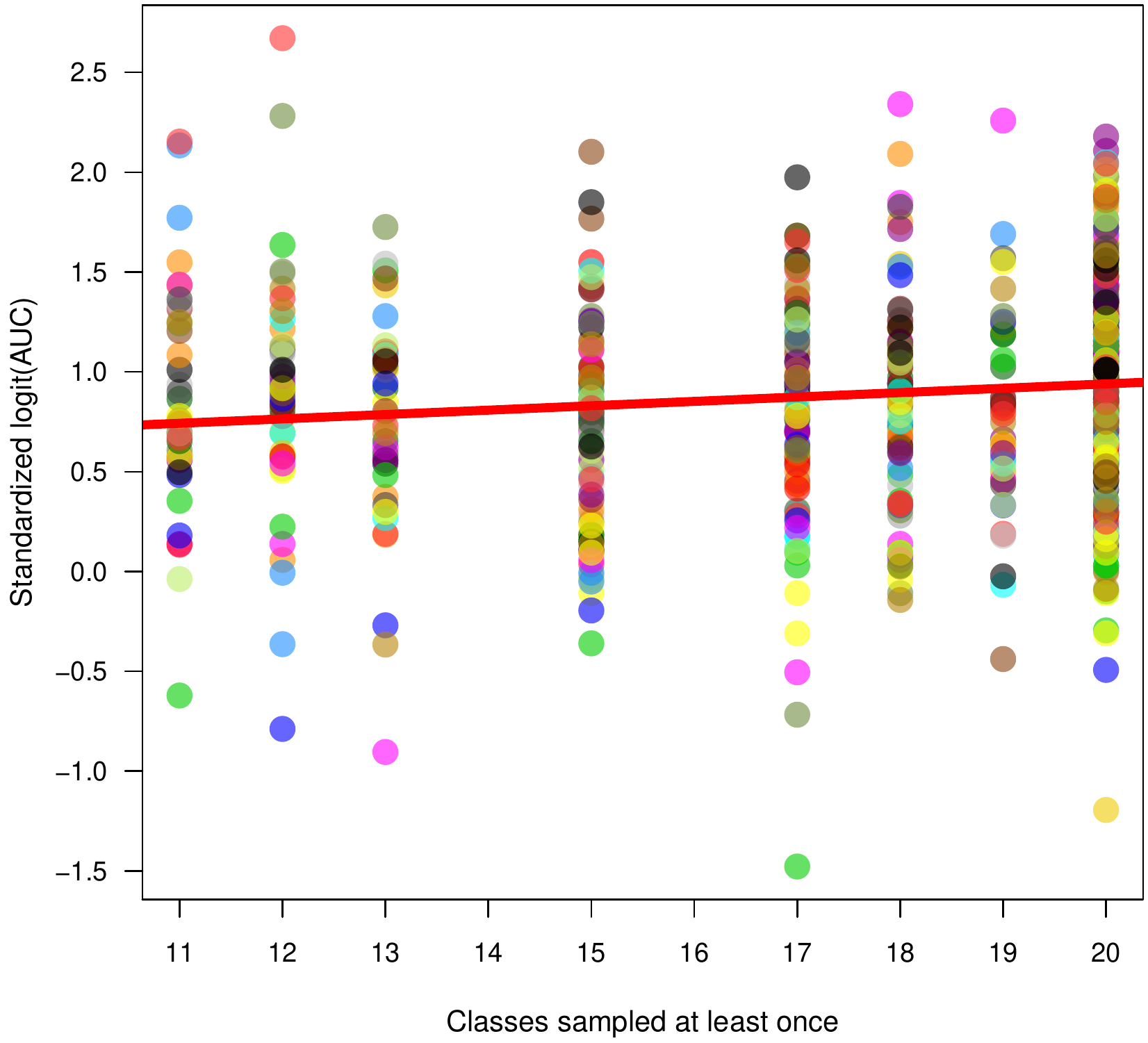} \\
            (e) 1024$\rightarrow$$\alpha$=0.019 &
            (f) 2048$\rightarrow$$\alpha$=0.007 &
            (g) 4096$\rightarrow$$\alpha$=0.011 &
            (h) 8192$\rightarrow$$\alpha$=0.022 \\
                (p-value=0.030) &
                (p-value=0.386) &
                (p-value=0.106) &
                (p-value=0.001) \\
        \end{tabular}
        \end{tiny}
     \end{center}
     \caption{Similar graphs of Figure~\ref{fig_bovw_plot_aps_vs_semantics_all} showing that even when we analyze the BoVW results per dictionary, we can see a very low impact of semantics in the precision. The first row has the plots versus semantic diversity and, the second row, versus the number of classes in each sample.}
     \label{fig_bovw_plot_aps_vs_semantics_each}
\end{figure}

    Figures~\ref{fig_bovw_plot_aps_vs_semantics_all} and~\ref{fig_bovw_plot_aps_vs_semantics_each} show the results for BoVW.
    In Figure~\ref{fig_bovw_plot_aps_vs_semantics_all}, all the dictionary configurations are shown together while in Figure~\ref{fig_bovw_plot_aps_vs_semantics_each}, each dictionary is shown separately.
    We can see that, as semantics increase in the dictionary (i.e., more classes are considered during dictionary creation), there is almost no increase in average precision.
    This is numerically visible by the very small values of the red line slope $\alpha$.
    We can also attest with high confidence (small p-value) that semantics has low importance for the dictionary quality.

\begin{figure}[t!]
     \begin{center}
      \scriptsize
        \begin{tabular}{@{}c@{ }c@{}}
            \includegraphics[width=0.45\columnwidth]{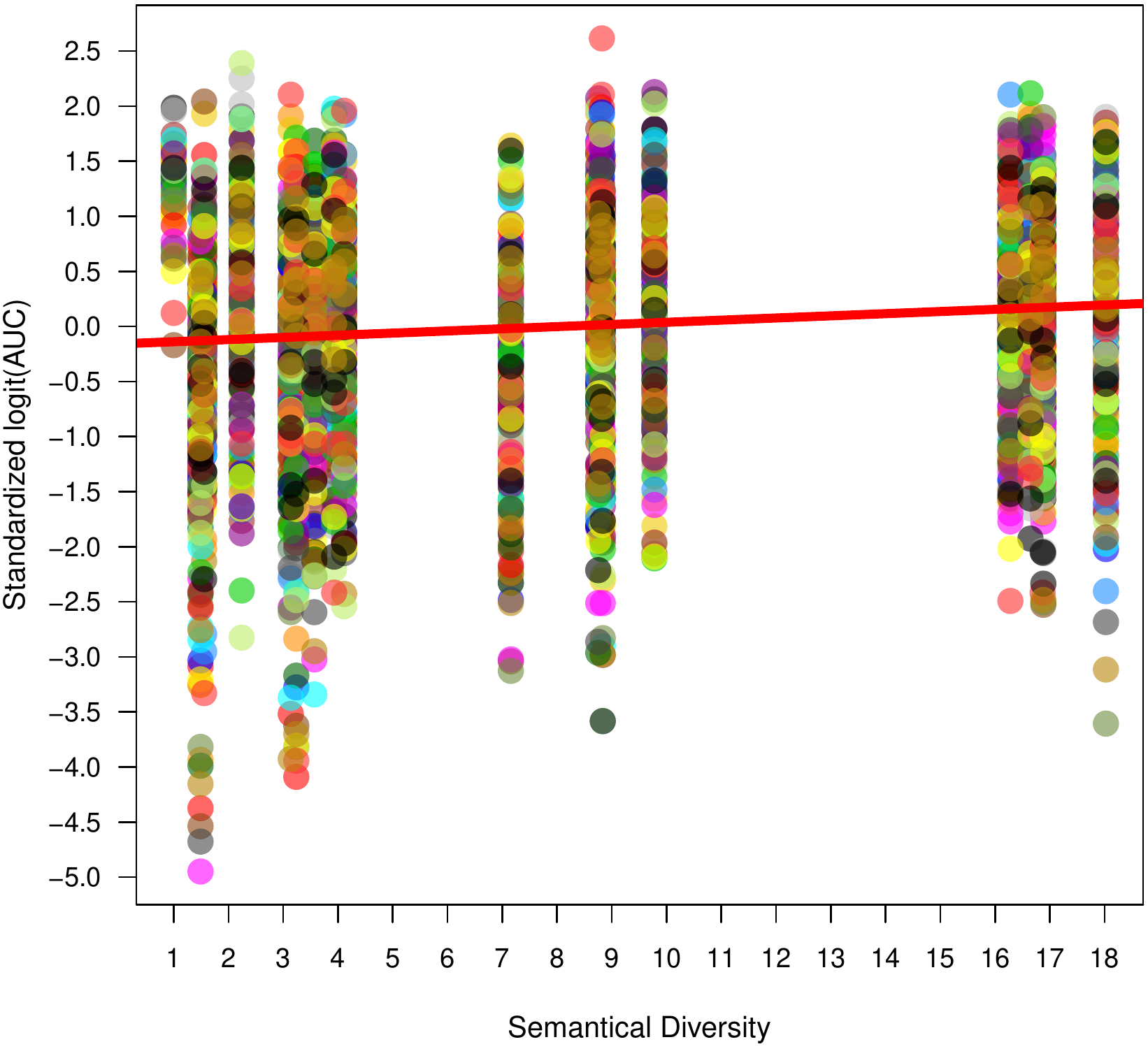} \hspace{0.5cm} &
            \includegraphics[width=0.45\columnwidth]{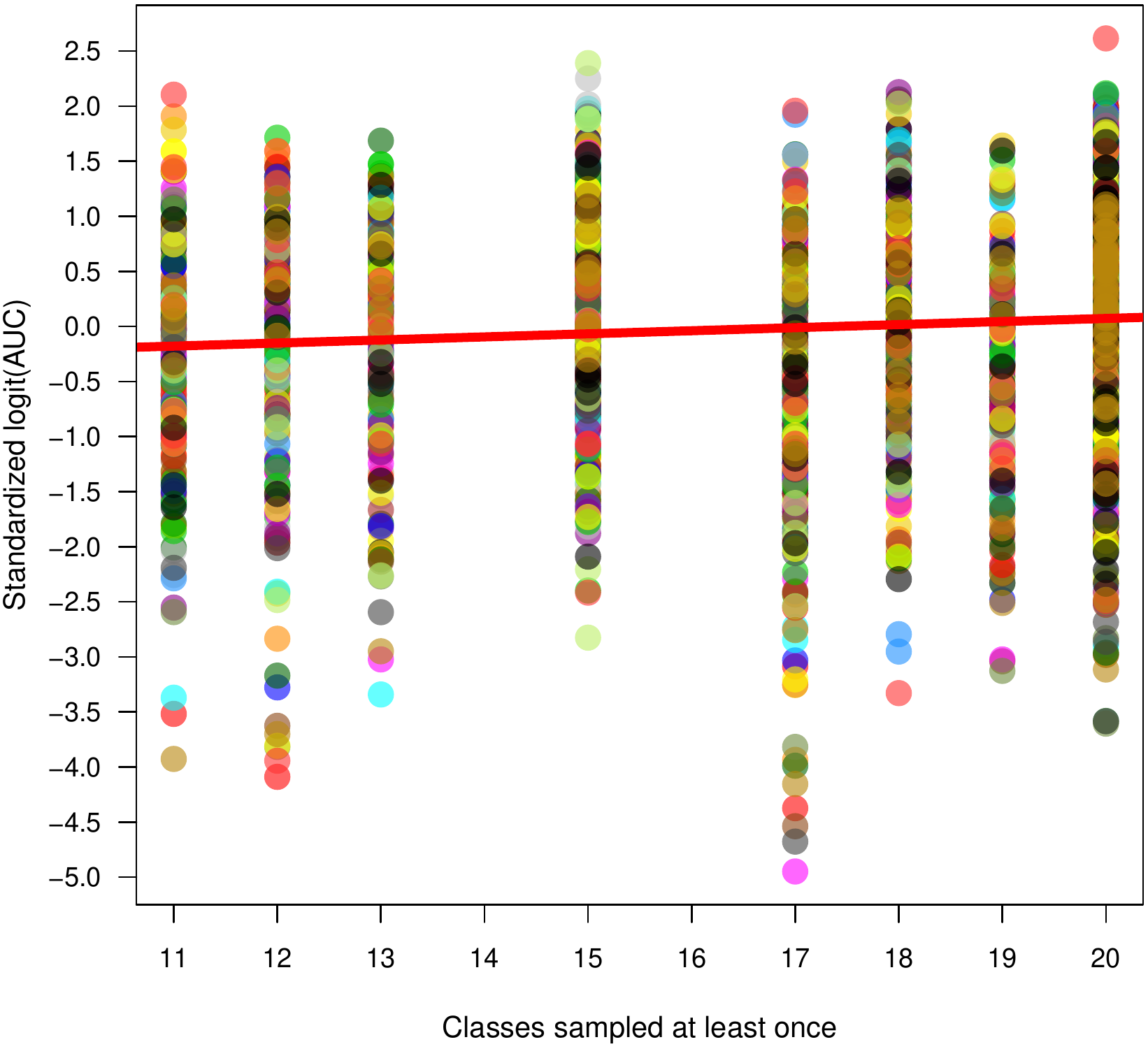} \\
            (a) $\alpha$ = 0.0196 (p-value=8.3e-12) &
            (b) $\alpha$ = 0.0279 (p-value=8.4e-7) \\
        \end{tabular}
     \end{center}
     \caption{For Fisher Vectors, similarly to BoVW (see Figure~\ref{fig_bovw_plot_aps_vs_semantics_all}), we can affirm with high confidence (very small p-value) that semantics has a very low impact on the codebook quality (coefficient $\alpha$ is very small).}
     \label{fig_fisher_plot_aps_vs_semantics_all}
\end{figure}

\begin{figure}[t!]
     \begin{center}
      \begin{tiny}
        \begin{tabular}{@{}c@{}c@{}c@{}c@{}c@{}}
            \includegraphics[width=0.20\columnwidth]{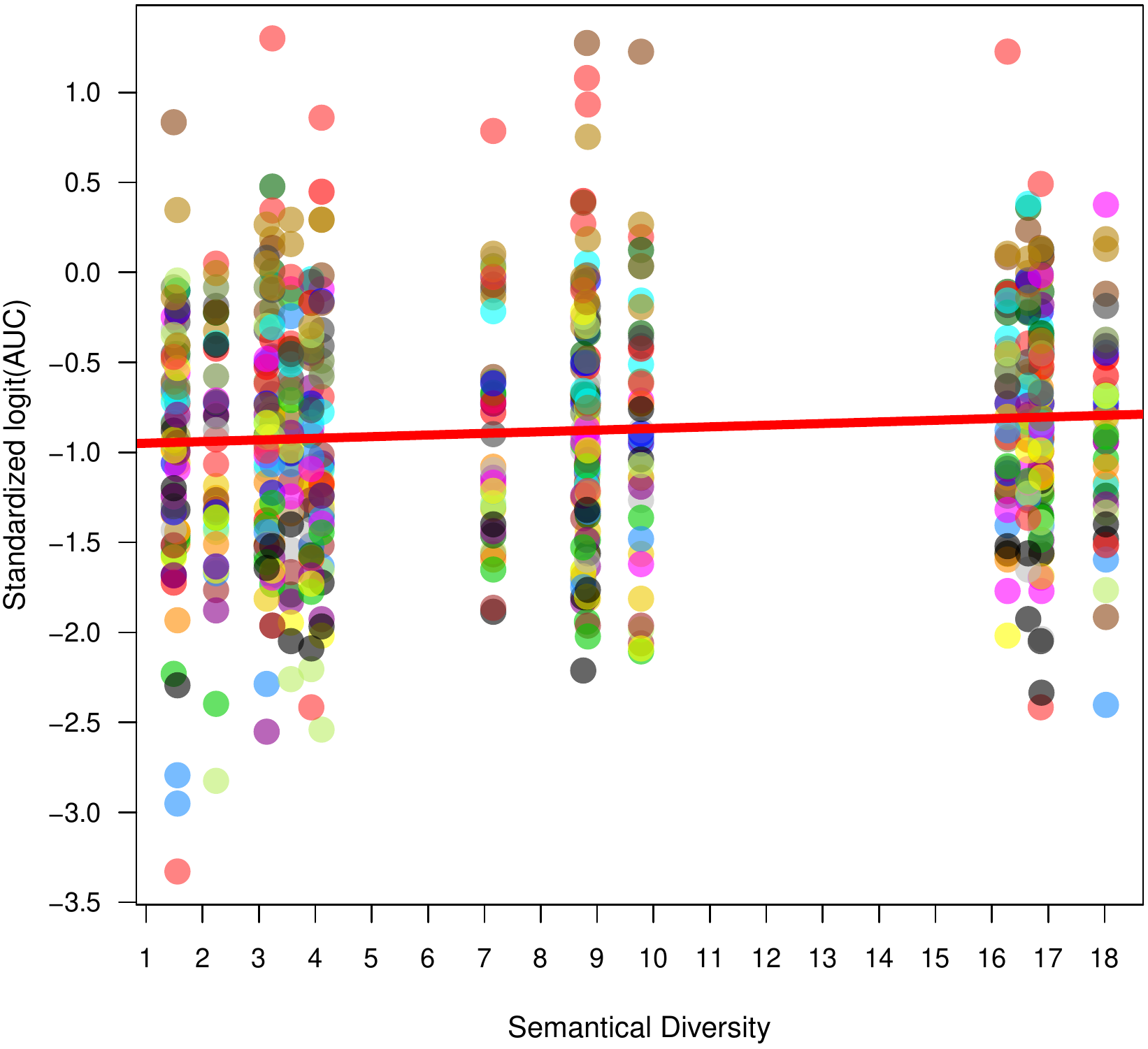} &
            \includegraphics[width=0.20\columnwidth]{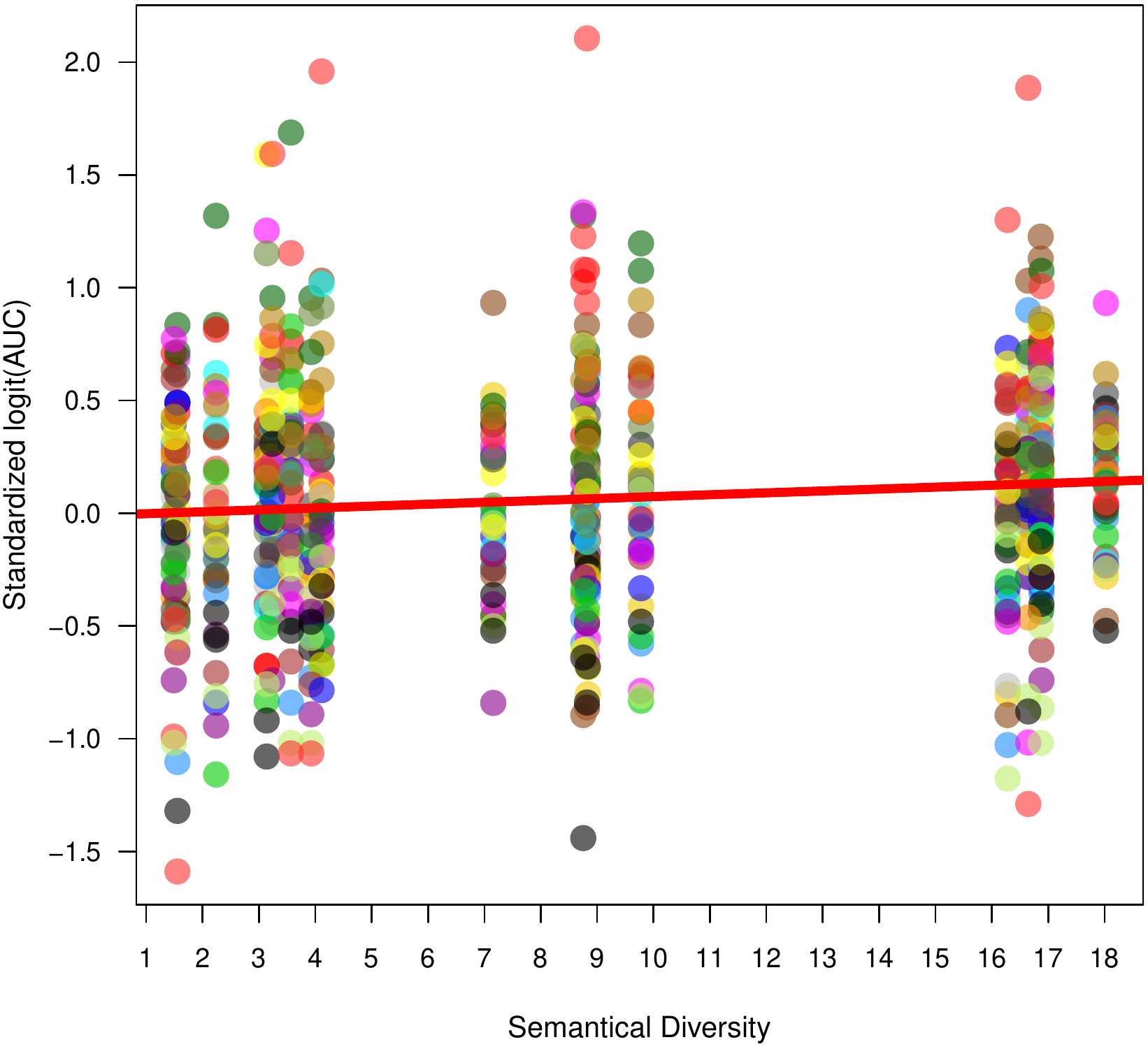} &
            \includegraphics[width=0.20\columnwidth]{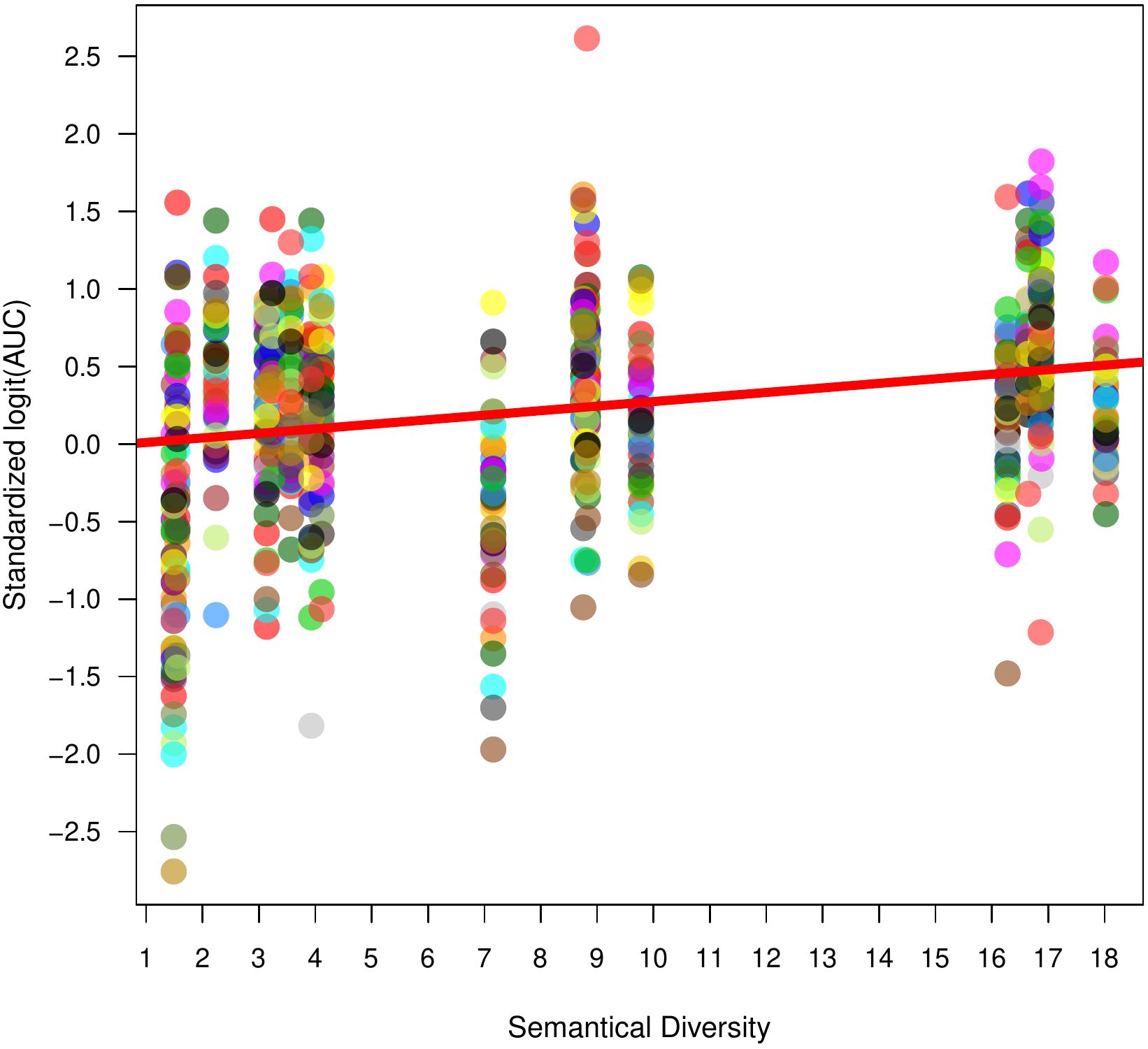} &
            \includegraphics[width=0.20\columnwidth]{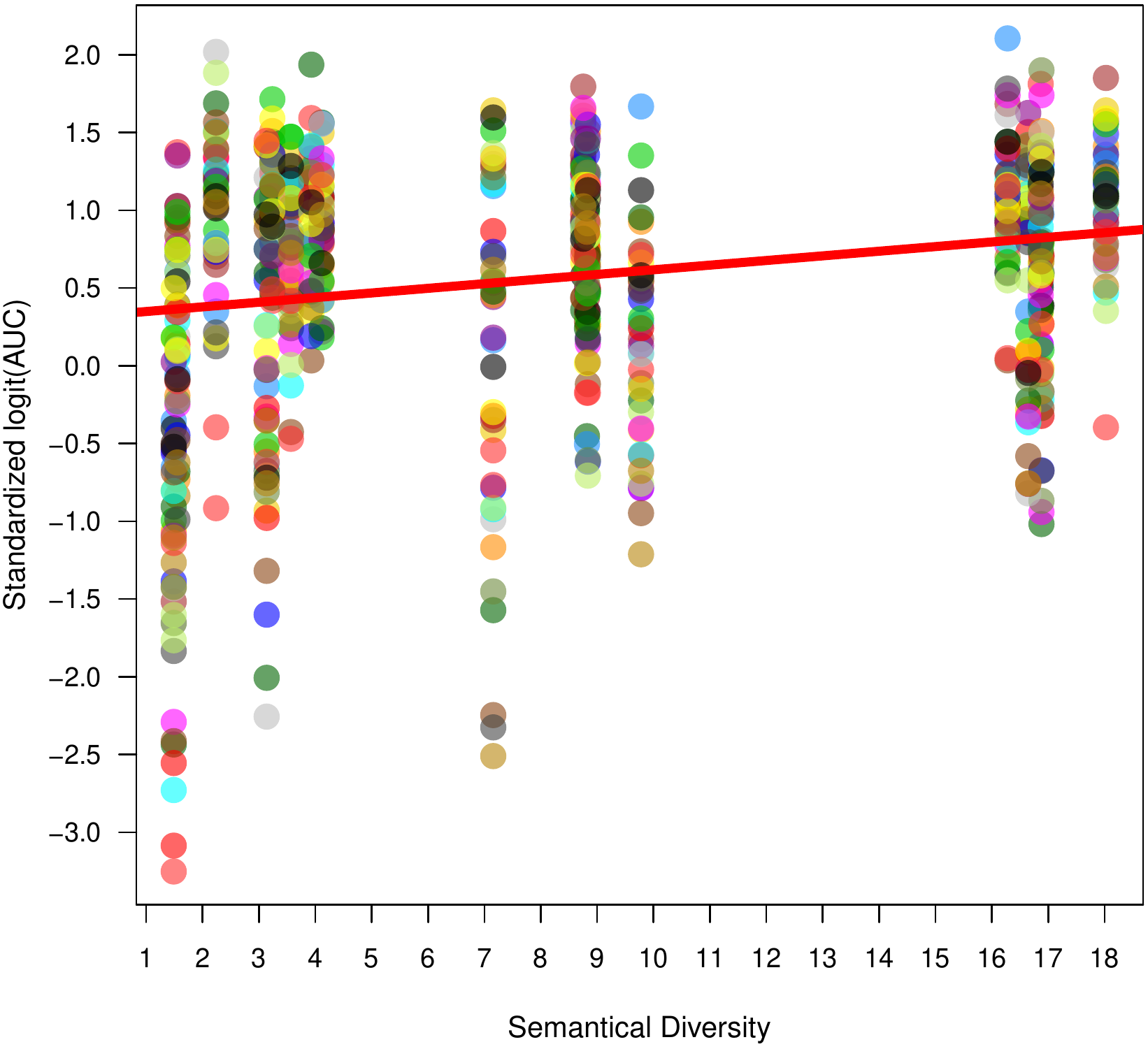} &
            \includegraphics[width=0.20\columnwidth]{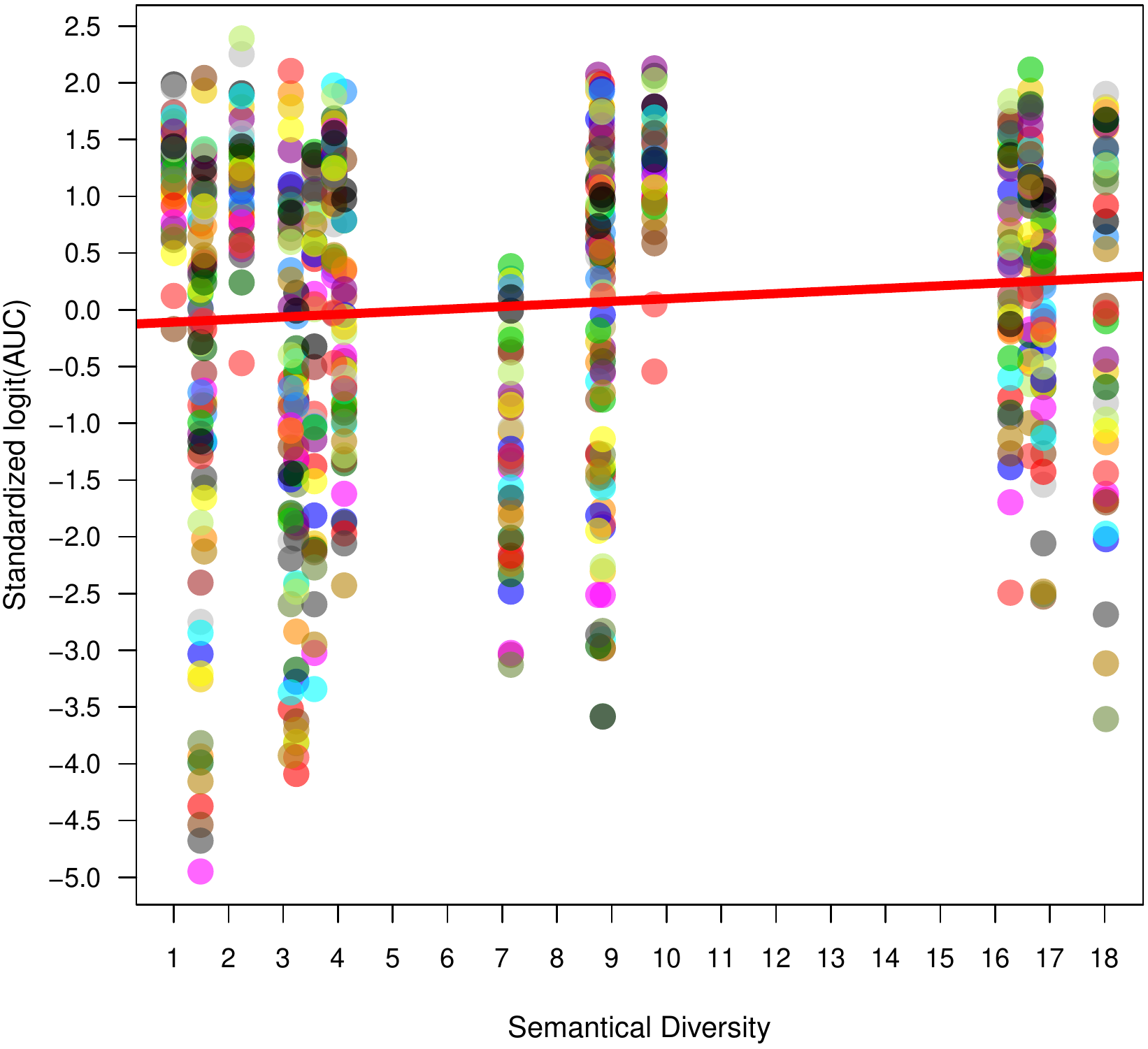} \\
            (a) 128$\rightarrow$$\alpha$=0.009 &
            (b) 256$\rightarrow$$\alpha$=0.008 &
            (c) 512$\rightarrow$$\alpha$=0.029 &
            (d) 1024$\rightarrow$$\alpha$=0.030 &
            (e) 2048$\rightarrow$$\alpha$=0.023 \\
                (p-value=0.029) &
                (p-value=0.008) &
                (p-value=3.9e-13) &
                (p-value=1.4e-8) &
                (p-value=0.016) \\

            \includegraphics[width=0.20\columnwidth]{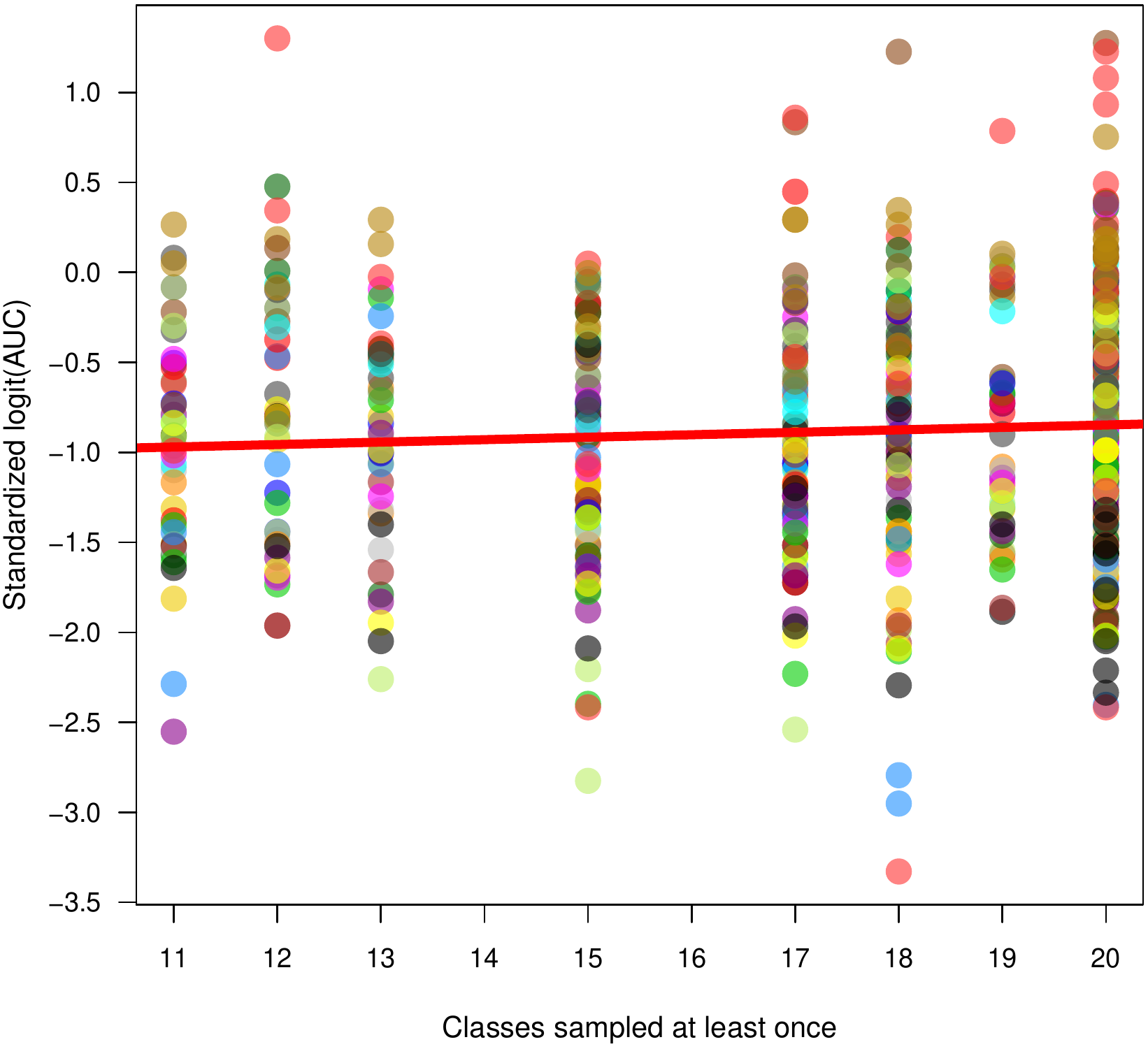} &
            \includegraphics[width=0.20\columnwidth]{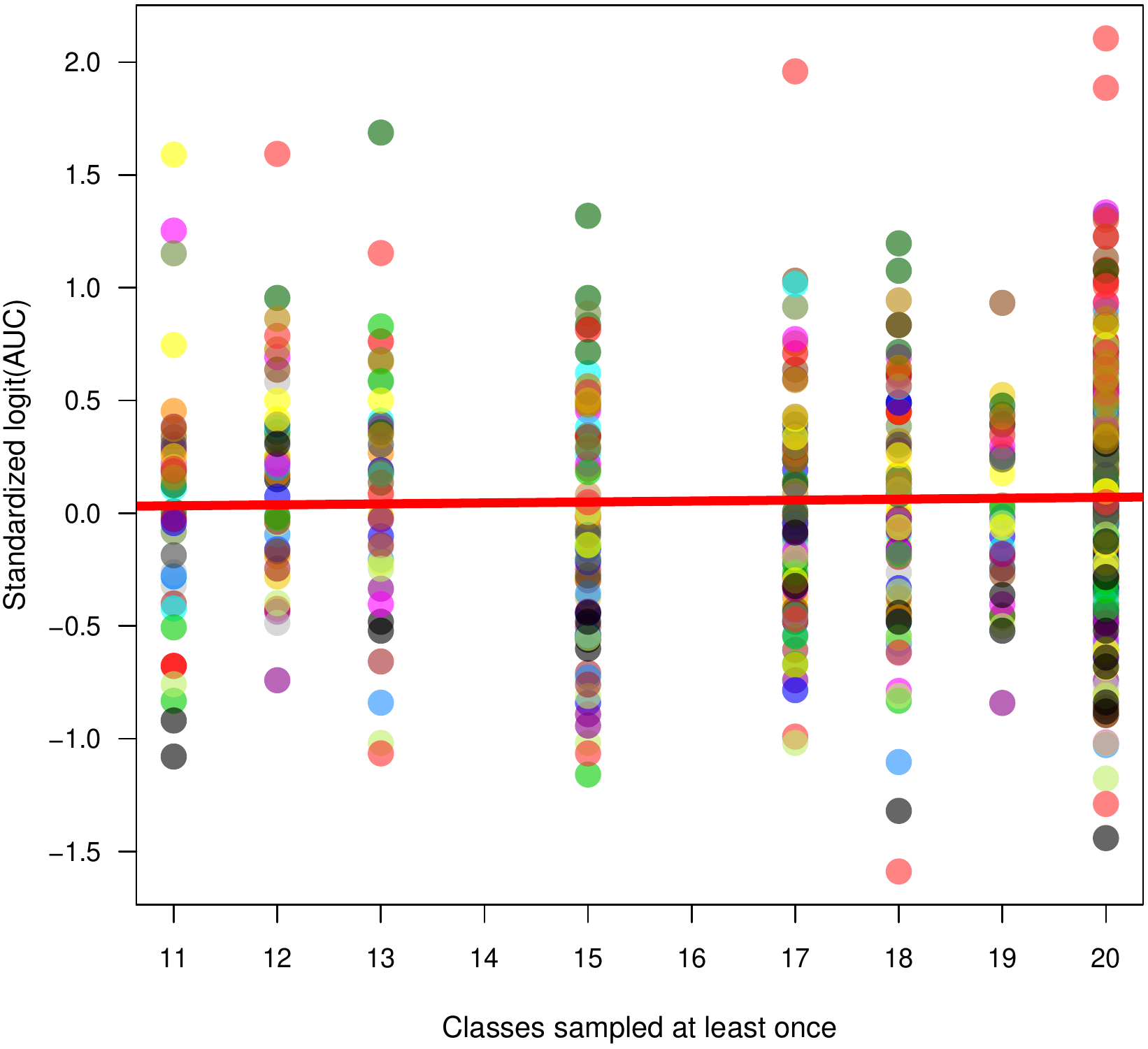} &
            \includegraphics[width=0.20\columnwidth]{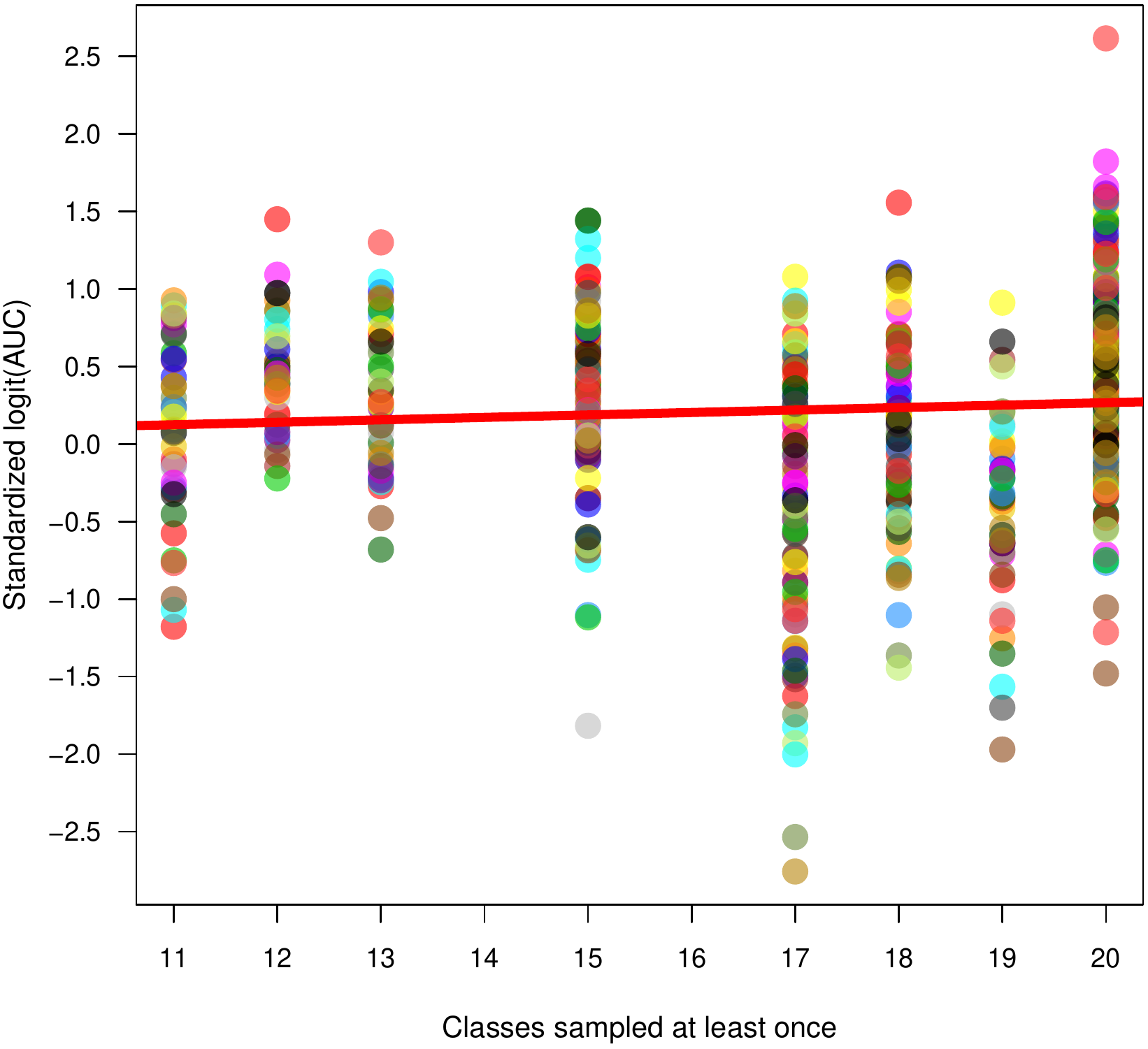} &
            \includegraphics[width=0.20\columnwidth]{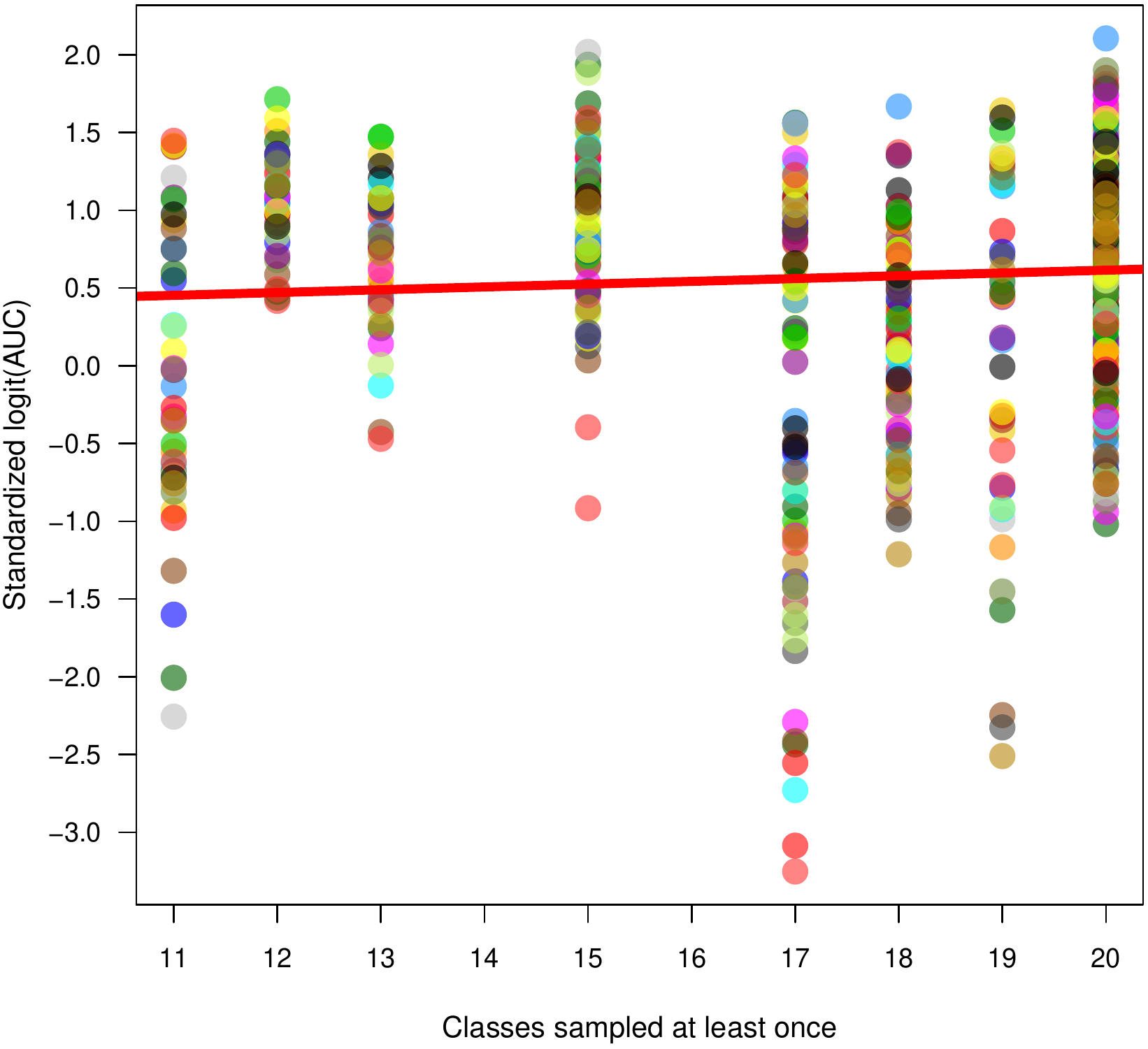} &
            \includegraphics[width=0.20\columnwidth]{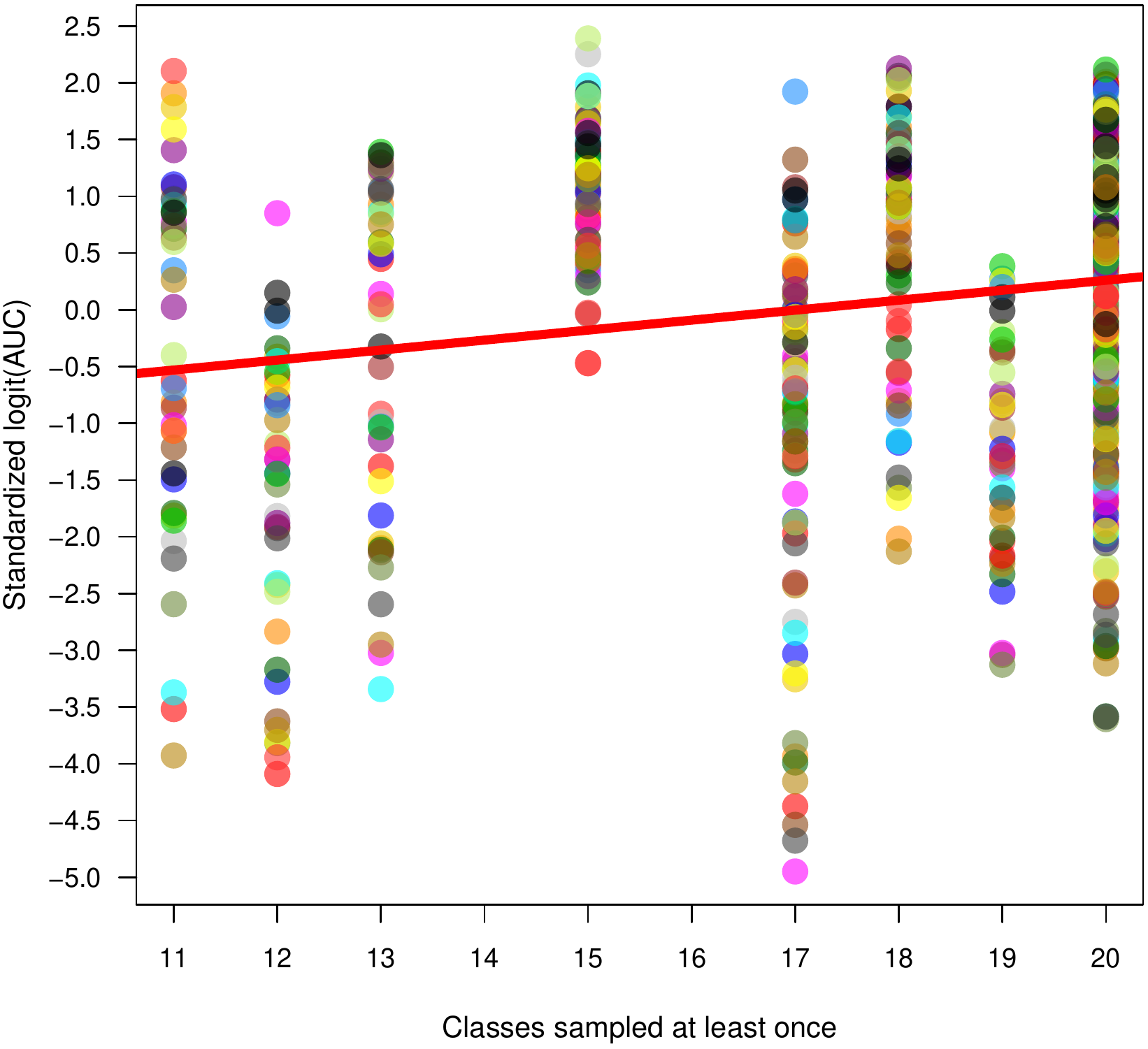} \\
            (f) 128$\rightarrow$$\alpha$=0.014 &
            (g) 256$\rightarrow$$\alpha$=0.004 &
            (h) 512$\rightarrow$$\alpha$=0.016 &
            (i) 1024$\rightarrow$$\alpha$=0.018 &
            (j) 2048$\rightarrow$$\alpha$=0.088 \\
                (p-value=0.099) &
                (p-value=0.507) &
                (p-value=0.052) &
                (p-value=0.087) &
                (p-value=1.5e-6) \\
        \end{tabular}
        \end{tiny}
     \end{center}
     \caption{Analogously to Figure~\ref{fig_bovw_plot_aps_vs_semantics_each}, Fisher Vectors have low variation in accuracy when varying the amount of semantics used for creating the dictionaries.}
     \label{fig_fisher_plot_aps_vs_semantics_each}
\end{figure}

    For Fisher Vectors, the results are presented in Figures~\ref{fig_fisher_plot_aps_vs_semantics_all} and~\ref{fig_fisher_plot_aps_vs_semantics_each}.
    Similarly to BoVW, semantics has few impact in the accuracy of Fisher Vectors.
    Although in some specific configurations, i.e., dictionaries of 1024 and 2048 (Figures~\ref{fig_fisher_plot_aps_vs_semantics_each}-\textit{d} and \textit{j}) the regression curve has a larger slope, its coefficient is still very small.

    Therefore, we can conclude that the impact of semantics in the dictionary quality is very low.
    Although the samples used for creating the dictionary can still be very poor in terms of semantics (few classes), they can be rich in terms of visual diversity, which is enough for creating a good dictionary.
    As the low-level descriptor (SIFT, in our experiments) captures image local properties and not semantics, the fast dictionary generalization occurs if we have a set of images rich enough in terms of textures, which will cover a good portion of the feature space without requiring the use of all image classes.
    We can also conclude that the dictionary size and the coding/pooling have a minor impact in those conclusions, as the observations above were possible regardless of the dictionary size and regardless of the coding/pooling method.

\section{Discussion}
\label{conclusions}

    In this paper, we evaluate the impact of semantic diversity in the quality of visual dictionaries.
    The experiments conducted show that dictionaries based on a subset of a collection may still provide good performance, provided that the selected sample is visually diverse.
    We showed experimentally that the impact of semantics in the dictionary quality is very small: classification results of dictionaries based on a sample with low or high semantic variability were similar.
    Therefore, we can point out that visual variability is more important than semantic variability when selecting the sample to be used as source of features for the dictionary creation.

    The scenarios that can benefit from our conclusions are the ones in which the dataset is dynamic (images are inserted and removed during the application use) and/or in which one do not have access to the whole dataset when the dictionary is created.
    Those scenarios are important because they reflect any application dealing with web-like collections.


    As a general recommendation for creating visual dictionaries in dynamic environments, we say that one must have a diverse set of images in terms of visual appearances and not necessarily in terms of classes. 
    Those findings open the opportunity to greatly alleviate the burden in generating the dictionary, since, at least for general-purpose datasets, we show that the dictionaries do not have to take into account the entire collection, and may even be based on a small collection of visually diverse images.

    In future work, we would like to explore if those findings also work on special-purpose datasets, like in quality control of industry images or in medical images, in which we suspect that a specific dictionary might have a stronger effect.
    We also would like to investigate possibilities for visualizing the codebooks, aiming at better understanding how different dictionaries cover the feature space.

\section*{Acknowledgment}
\label{ack}
    Thanks to Fapesp (grant number 2009/10554-8), CAPES, CNPq, Microsoft Research, AMD, and Samsung.

\ifCLASSOPTIONcaptionsoff
  \newpage
\fi



%

\bibliographystyle{IEEEtran}
\bibliography{crossDictionaries}

%








\end{document}